\documentclass[12pt]{article}

\usepackage{scicite}

\usepackage{times}

\usepackage{graphicx}
\usepackage{gensymb}
\usepackage{amsmath}
\newenvironment{sciabstract}{%
\begin{quote} \bf}
{\end{quote}}

\title{Impact-resistant, autonomous robots inspired by tensegrity architecture}

 \author
{William R. Johnson III,$^{1\dag}$ Xiaonan Huang,$^{1\dag}$ Shiyang Lu,$^{2}$ Kun Wang,$^{2}$ \\Joran W. Booth,$^{1}$ Kostas Bekris,$^{2}$ Rebecca Kramer-Bottiglio$^{1\ast}$\\
\\
\normalsize{$^{1}$Mechanical Engineering and Materials Science, School of Engineering and Applied Science}\\
\normalsize{Yale University, New Haven, CT 06520 USA}\\
\normalsize{$^{2}$Computer Science Department, Rutgers University, Piscataway, NJ 08854 USA}\\
\\
\normalsize{$^\dag$These authors contributed equally to this work.}\\
\normalsize{$^\ast$Corresponding author. E-mail:  rebecca.kramer@yale.edu.}
}

\date{}

\begin{document} 

% Double-space the manuscript.

\baselineskip24pt

% Make the title.

\maketitle

% Place your abstract within the special {sciabstract} environment.

\begin{sciabstract}
  % Future robots will navigate perilous, remote environments with resilience and autonomy.  Researchers have proposed building robots with compliant bodies to enhance robustness, but this approach often sacrifices the autonomous capabilities expected of rigid robots.  Inspired by tensegrity architecture, we introduce a tensegrity robot---a hybrid robot made from rigid struts and elastic tendons---that demonstrates the advantages of compliance and the autonomy necessary for robotic task performance.  This mobile robot platform demonstrates impact resistance and autonomy in a field environment and boasts additional advances in the state of the art, including surviving harsh impacts from drops (at least 5.7~m), accurately reconstructing its shape and orientation using on-board sensors, achieving high locomotion speeds (18~bar lengths per minute), and climbing the steepest incline of any tensegrity robot (28~degrees).  We characterize the robot's locomotion on unstructured terrain, showcase its autonomous capabilities in a trajectory following task, exhibit its shape morphing in a game of limbo, and demonstrate its robustness by rolling it off a cliff.

  Future robots will navigate perilous, remote environments with resilience and autonomy.  Researchers have proposed building robots with compliant bodies to enhance robustness, but this approach often sacrifices the autonomous capabilities expected of rigid robots.  Inspired by tensegrity architecture, we introduce a tensegrity robot---a hybrid robot made from rigid struts and elastic tendons---that demonstrates the advantages of compliance and the autonomy necessary for task performance.  This robot boasts impact resistance and autonomy in a field environment and additional advances in the state of the art, including surviving harsh impacts from drops (at least 5.7~m), accurately reconstructing its shape and orientation using on-board sensors, achieving high locomotion speeds (18~bar lengths per minute), and climbing the steepest incline of any tensegrity robot (28~degrees).  We characterize the robot's locomotion on unstructured terrain, showcase its autonomous capabilities in navigation tasks, and demonstrate its robustness by rolling it off a cliff.
  
  % from APS abstract
  % Highly stretchable sensors for feedback control allow this robot to achieve high speeds (34 body lengths per minute), climb the steepest incline (28 degrees) of any tensegrity robot, navigate unstructured terrains (pebbles, grass, sand, etc.), and accurately reconstruct its shape and orientation (within 10\% of the bar length) relative to Earth's gravitational and magnetic fields.
\end{sciabstract}

% In setting up this template for *Science* papers, we've used both
% the \section* command and the \paragraph* command for topical
% divisions.  Which you use will of course depend on the type of paper
% you're writing.  Review Articles tend to have displayed headings, for
% which \section* is more appropriate; Research Articles, when they have
% formal topical divisions at all, tend to signal them with bold text
% that runs into the paragraph, for which \paragraph* is the right
% choice.  Either way, use the asterisk (*) modifier, as shown, to
% suppress numbering.

\section*{Introduction}

In the mid-20th century, art student Kenneth Snelson and his professor, architect Buckminster Fuller, began creating peculiar sculptures by suspending rigid elements in a tension network made from strings~\cite{snelson1996snelson}.  These ``discontinuous compression'' sculptures had useful properties, including flexibility, light weight, minimal material usage, high strength-to-weight ratios, and the ability to absorb large external loads~\cite{snelson1965continuous}.  Fuller would later name this concept {\em tensegrity}, a portmanteau of ``tensile integrity''~\cite{micheletti2022seventy}.  Since its invention, the principle of tensegrity has been applied to numerous fields, including art~\cite{snelson2012art}, architecture~\cite{buckminster1962tensile}, civil engineering~\cite{gilewski2015applications}, biomechanical modeling~\cite{chen1999tensegrity,wang2001mechanical}, and robotics~\cite{shah2021tensegrity,liu2022review}.

Made from rigid struts and elastic tendons, tensegrity robots have the potential to navigate perilous and unpredictable environments.  Their inherently compliant bodies make them adaptable, deployable, and resilient~\cite{skelton2001introduction}.  These advantages, along with their light weight, make tensegrity robots a promising candidate for the next generation of planetary rovers~\cite{sabelhaus2015system} and for navigating other remote environments.  In recent years, researchers have introduced locomotive tensegrity robots that are capable of crawling~\cite{paul2006design,zappetti2017bio,kobayashi2022soft}, rolling~\cite{chen2017soft,vespignani2018design,wang2019light,kinodynamic_tensegrity,chen2017inclined,kaufhold2017indoor,rhodes2019compact,baines2020rolling,kim2020rolling,booth2020surface,surovik2021adaptive}, vibrating~\cite{bohm2013vibration,rieffel2018adaptive,kimber2019low}, jumping~\cite{chung2019jumping,garanger2020soft}, climbing~\cite{friesen2016second,kobayashi2022soft}, swimming~\cite{chen2019swimming,shintake2020bio}, and flying~\cite{mintchev2018soft,zha2020collision}.  Tensegrity robots achieve rolling, the most common locomotion mode, by deforming their bodies to shift their center of mass outside of the stable polygon formed by the nodes in contact with the ground, using either electric motors~\cite{vespignani2018design} or other actuators~\cite{wang2019light,baines2020rolling} to change the lengths of their tendons or bars. 

% These initial studies and hardware demonstrations of tensegrity robots show promise, but they also leave much to be desired.

% Autonomous tensegrity robots navigating unstructured environments have been precluded by unsolved challenges in design, sensing, and control.
Despite their exciting advantages, tensegrity robots are not yet able to autonomously navigate unstructured environments due to unsolved problems in design, sensing, modeling, and control.
% To reach their full potential navigating unstructured and challenging environments,
To solve this grand challenge,
tensegrity robots need reliable sensors for state estimation and control, increased levels of autonomy, versatility across different terrains, and demonstrated resilience to harsh impacts~\cite{shah2021tensegrity,gao2017review}.  There has been some work on state estimation for tensegrity robots~\cite{caluwaerts2016state,lu20226n}, including shape reconstruction with on-board sensors~\cite{booth2020surface,li2021shape}.  To prove their adaptability and versatility, some researchers demonstrate tensegrity robots on unstructured terrain~\cite{chen2017soft,wang2019light,vespignani2018design,kobayashi2022soft}.  Tensegrity robots have been shown to climb inclines as steep as 24~\degree~\cite{chen2017inclined}.  There are also a few examples of tensegrity robots exhibiting robustness to impacts by surviving short drops~\cite{rieffel2018adaptive,vespignani2018design,zappetti2022dual,spiegel2023shape}.  Despite these advancements, an autonomous tensegrity robot with impact resistance has yet to be demonstrated.

In this work, we introduce Tribar, a 3-bar tensegrity robot, and demonstrate its capabilities of state estimation, locomotion, autonomous control, shape morphing, and impact resistance.  The robot is capable of extreme deformations, and therefore we use highly stretchable strain sensors~\cite{johnson2022sensor} for state estimation and closed-loop control.  Beyond what has been demonstrated previously, this tensegrity robot can estimate its shape and global orientation in real time with on-board sensors, enabling uninterrupted autonomous locomotion in a field environment after a 2~m fall (Fig.~\ref{fig:splash}).
We characterize the robot's locomotion across different unstructured terrains, with different gaits, and up inclines as steep as 28\degree, which is the steepest climbed by a tensegrity robot in the reported literature. Autonomous control is further demonstrated in trajectory following experiments that make use of the tensegrity's multiple gaits for forward locomotion and turning.  In a game of limbo, the robot exhibits autonomy and shape morphing, which allow it to navigate under an obstacle with decreasing height.  Finally, we show the robot rolling off a bridge and surviving a 5.7~m drop, the highest in the reported literature.  This work represents an advance in the state-of-the-art across the quantitative metrics pertinent to tensegrity robots [supplementary materials (SM) section S1] and a step toward realizing tensegrity robots' promise to autonomously explore unstructured and perilous remote environments.

\section*{Design of a 3-bar Tensegrity Robot}

% figure one
% panel A: locomotion in field environment
% panel B: Design of robot (labeled diagram)
% panel C: Closed-loop control (locomotion trial and sensor/tracking plot) highest speed? dynamic locomotion?

\begin{figure}
    \centering
    \makebox[\textwidth][c]{\includegraphics[width=7.24in]{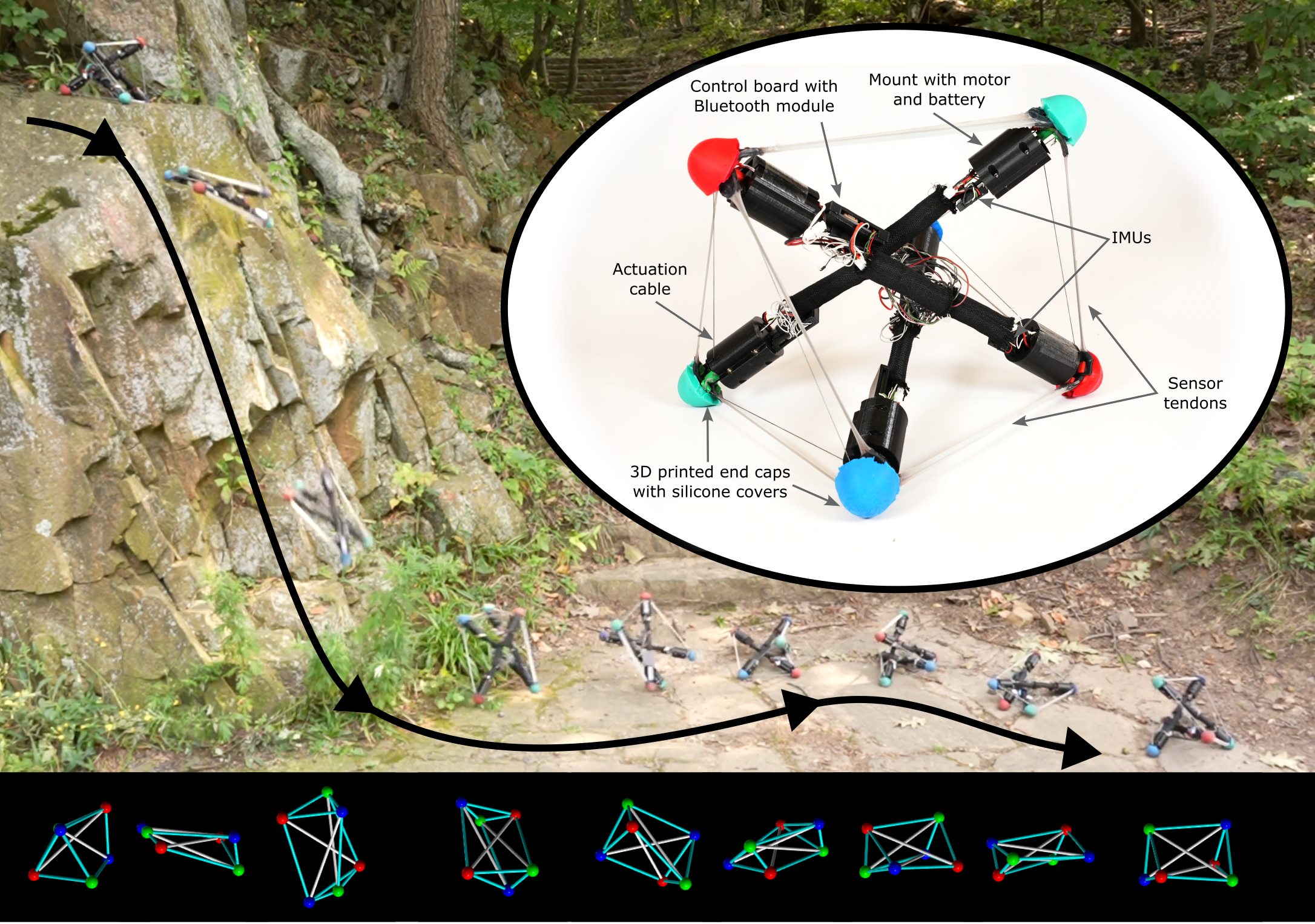}}
    \caption{\textbf{An impact-resistant tensegrity robot exhibits autonomous locomotion in an unstructured environment.}  A 3-bar tensegrity robot rolls off a 2~m cliff and survives the crash landing due to its inherent body compliance.  Afterward, it continues its locomotion in the same direction.  The robot uses on-board sensors to estimate its shape and orientation.  Estimating its cable lengths, downward face, and heading allows the robot to continue its locomotion after the disturbance caused by the initial fall.
    % The sensor information feeds back into the autonomous controller, which maintains the robot's locomotion 
    % cable lengths, downward face, and heading. allowing it to continue locomotion autonomously after the disturbance caused by the initial fall.
    % (A) A 3-bar tensegrity robot exhibiting rolling locomotion in a field environment. (B) The robot can climb a maximum incline of 28 degrees, the steepest climbed by a tensegrity robot in the published literature. (C) Demonstrating notable impact resistance, the robot rolls off a 5.7~m bridge and survives the drop. (D) The robot comprises three carbon fiber rods connected by cables and elastic strain sensors.  An onboard microcontroller and motor drivers control the six actuators, DC motors that drive winches to extend and contract cables.  Two onboard IMUs measure the orientation of the robot.
    % (C) The robot achieves locomotion through extreme deformations that shift its center of gravity outside of its polygon of stability.  Elastic strain sensors are used with feedback control to ensure the robot reliably executes a gait, comprising a sequence of target shapes.
    }
    \label{fig:splash}
\end{figure}

% design of the robot
%%% explanation of structural components
%%% description of onboard electronics
%%% capacitive strain sensors

Tribar is made from three rigid struts connected with tensile elements in a stable network (Fig.~\ref{fig:splash}).  Each 360~mm strut contains two motors that can extend and contract cables, deforming the tensegrity structure.  The six short tendons are actuated by the six motors, while the longer three tendons act as passive restoring springs.  The robot is untethered and carries its own power in the form of six 1S LiPo batteries, two on each strut, wired 3S2P for a total of 11.1~V nominal.  The electronics are distributed among the three struts, including a microcontroller with Bluetooth module, motor drivers, inertial measurement units (IMUs), and a capacitive sensing unit.  Each strut contains custom 3D printed parts, including end caps (Fig.~\ref{fig:endcaps}) and mounts for electronics.  Full details are found in SM section M1, and a circuit schematic is shown in Fig.~\ref{fig:electronics}.

Each connection between the struts has one highly stretchable capacitive strain sensor that doubles as a tensegrity tendon~\cite{johnson2022sensor}. These sensor tendons (Fig.~\ref{fig:sensors}) are made from two thin layers of a room-temperature liquid metal paste~\cite{wang2019printed,kong2020oxide} encapsulated in a silicone elastomer.  The capacitance of the sensors varies linearly as they are stretched, and the sensors can accommodate the high strains ($>300\%$) that are necessary for tensegrity robot locomotion.  Fabrication details and characterization are given in SM section M1.3 and in our previous work~\cite{johnson2022sensor}.  In this work, the six sensor tendons that are parallel to the motor-driven cables are used for feedback control.  The other three sensor tendons are stiffer, and they are used as passive restoring springs as well as for shape estimation.

% robot locomotion
%%% low-level feedback control
%%% gaits are sequences of target shapes
%%% rolling locomotion via shifting COM
%%% handles unstructured terrain

With feedback from the sensor tendons, the robot can reliably achieve target shapes by extending and contracting the six actuated cables until all the sensor lengths match their target lengths within some small tolerance (SM section M3).  A gait comprises a repeating sequence of target shapes that achieve a desired behavior like rolling or turning.  A hand-tuned PID controller commands the motors to achieve target lengths, and once all six lengths have been reached, the robot proceeds to the next target shape in the gait.  During rolling gaits, the tensegrity robot deforms its body to shift its center of mass outside of its polygon of stability and topple onto an adjacent face.  Some of the robot's gaits were designed by hand while others were discovered in simulation~\cite{wang2022real2sim2real}.  The complete repertoire of gaits can be found in SM sections M3, S4, and S6.
% This gravity-based rolling gait is undeterred by unstructured terrain found in natural environments (Fig. 1A).
% The robot achieves its top speed of 33 bar lengths per minute (BLPM) with its dynamic rolling gait (see materials and methods).

%% I think we will discuss the significance of this top speed in the introduction

The tensegrity robot was designed for impact resistance and autonomous control, as shown in Fig. 1.  With closed-loop control, the robot executes a rolling gait and rolls off a cliff, falling 2~m.  After absorbing the impact with its inherent body compliance, the robot senses its shape as well as its orientation and continues its locomotion in the same direction.  While other autonomous mobile robots would have to plan a path around obstacles like cliffs, our tensegrity robot can just roll off the edge, demonstrating its potential for efficiently navigating unstructured environments. 

\section*{State Estimation}

% figure two
% panels of locomotion trial(s)
% plots

\begin{figure}
    \centering
    \makebox[\textwidth][c]{\includegraphics[width=7.24in]{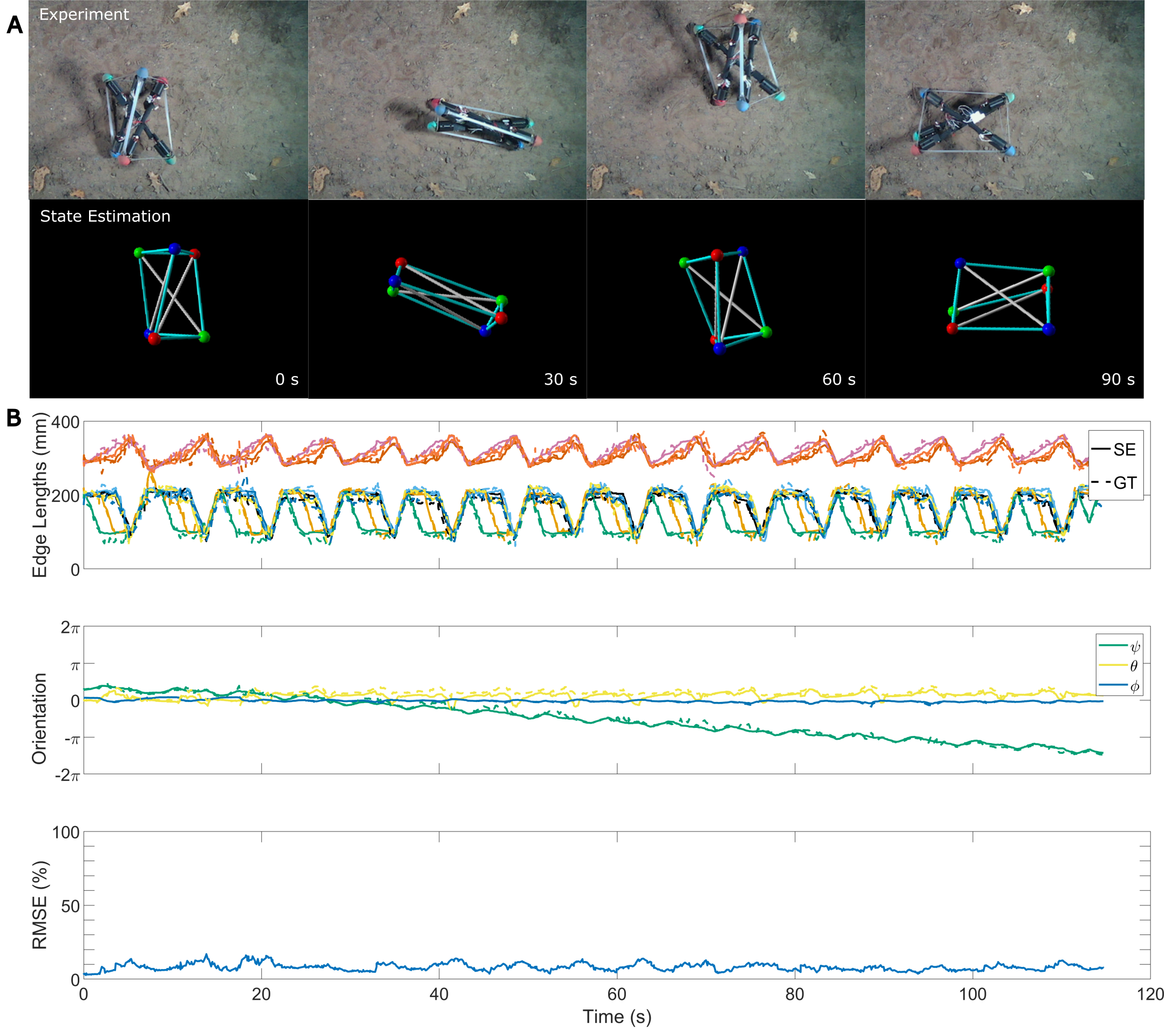}}
    \caption{\textbf{Real-time state estimation.} (A) A locomotion trial in a field setting with photographs and corresponding renderings of the robot's estimated state from onboard sensors. (B) A plot of the same locomotion trial showing the results of state estimation (SE) from onboard sensors compared to ground truth (GT) from our vision-based pose tracking algorithm~\cite{lu20226n}.  The top subplot shows the nine tendon lengths, the middle subplot shows the ZYX Euler angles that describe the robot's orientation relative to the global frame defined by Earth's gravitational and magnetic fields, and the bottom subplot shows the root mean square error (RMSE) between the estimated and ground truth positions of the nodes relative to the bar length.  The robot executes its counterclockwise turning gait.}
    \label{fig:state_recon}
\end{figure}

% importance of state reconstruction
%%% significance compared to the state of the art
%%% grand challenge

Robots that navigate unknown environments must be able to estimate their own state for locomotion and control.  Due to tensegrity robots' many degrees of freedom and coupled dynamics, state estimation remains a grand challenge~\cite{shah2021tensegrity}.  Previous work on tensegrity state estimation has relied on off-board sensors that realistically will not be available in unknown environments~\cite{caluwaerts2016state,lu20226n}.  Using only on-board sensors, researchers have been able to estimate the shape~\cite{li2021shape} or shape plus downward face~\cite{booth2020surface} of a tensegrity robot using stretchable sensors that strain $<100\%$.  Our 3-bar tensegrity robot uses highly stretchable sensor tendons and IMUs to reconstruct its shape and orientation in a global frame.
% relative to Earth's gravitational and magnetic fields.  
While previous work has been limited to stationary experiments (SM section S1), our state estimation runs in real time during locomotion (Fig.~\ref{fig:state_recon} and Fig.~\ref{fig:state_reconstruction2}).
% Further discussion on state estimation for tensegrity robots can be found in SUPPLEMENTARY MATERIALS.

% state reconstruction hardware and algorithm
%%% strain sensors and IMU
%%% constrained optimization
%%% results

% Fig. 2 shows the results of the robot's real-time state reconstruction.  First, the readings from the nine highly stretchable sensor tendons along with the known bar lengths are used to solve a constrained optimization that estimates the robot's shape in an arbitrary frame.  Then, the shape is rotated in a second optimization routine to minimize the error between the estimated bar orientations and the orientations measured by two onboard IMUs.  The robot reconstructs its state (shape plus global orientation) with an error less than 10\% of its bar length compared to a vision-based tensegrity pose tracking algorithm we developed in previous work~\cite{lu20226n}.  Moreover, the state reconstruction runs in real-time using only on-board sensors during locomotion, a feat possible because the highly stretchable sensors do not encumber the large demonstrations necessary for tensegrity locomotion.

In this work, the robot's state is defined as its shape and orientation.  In order to sense translation, the robot requires an external camera (SM section S4). The output of state estimation is the 3D position vector of each of the robot's six nodes centered at the origin and oriented relative to Earth's gravitational and magnetic fields. The inputs to the algorithm are the nine lengths measured by the sensor tendons and two bar orientations given by the digital motion processors on the 9-axis IMUs.  The estimation algorithm is a two-stage constrained optimization process.  First, the shape is estimated by minimizing the squared error from the measured tendon lengths constrained by the known bar lengths, as in previous work~\cite{booth2020surface}.  Second, the orientation is estimated by minimizing the error between the estimated and measured bar orientations.  Full details of the state estimation algorithm are given in SM section M2.

Fig.~\ref{fig:state_recon} shows the results of the robot's real-time state estimation while executing its counterclockwise turning gait on a dirt path.  Fig.~\ref{fig:state_recon}A shows a side-by-side comparison of video frames and state estimation results for the 2-minute trial.  The plots in Fig.~\ref{fig:state_recon}B shows the tendon lengths, orientation, and error as a function of time.  Dashed lines with corresponding colors indicate ground truth as measured by our vision-based tensegrity pose tracking algorithm~\cite{lu20226n}.  The orientation is plotted as the ZYX Euler angles that define the robot's orientation relative to the global frame, and the error is the root mean square error (RMSE) normalized by the robot's 360~mm bar length. The RMSE over the trial in Fig.~\ref{fig:state_recon} is 8.4\% of the bar length.  Another state reconstruction experiment with the quasistatic rolling gait is shown in Fig.~\ref{fig:state_reconstruction2}, and the RMSE from that trial is 10\%.

In practice, the state estimation algorithm is sufficiently fast and accurate for real-time autonomous control.  Before each step in the rolling gait of Fig.~\ref{fig:splash}, the robot inferred its downward face and heading from state estimation and adapted its gait via symmetry reduction (see SM section S6) to ensure it would continue rolling in the same direction it was facing at the start of the trial.

\section*{Locomotion}

% figure three
% photoframes and bar graphs of multi-terrain and incline locomotion

\begin{figure}
    \centering
    \makebox[\textwidth][c]{\includegraphics[width=7.24in]{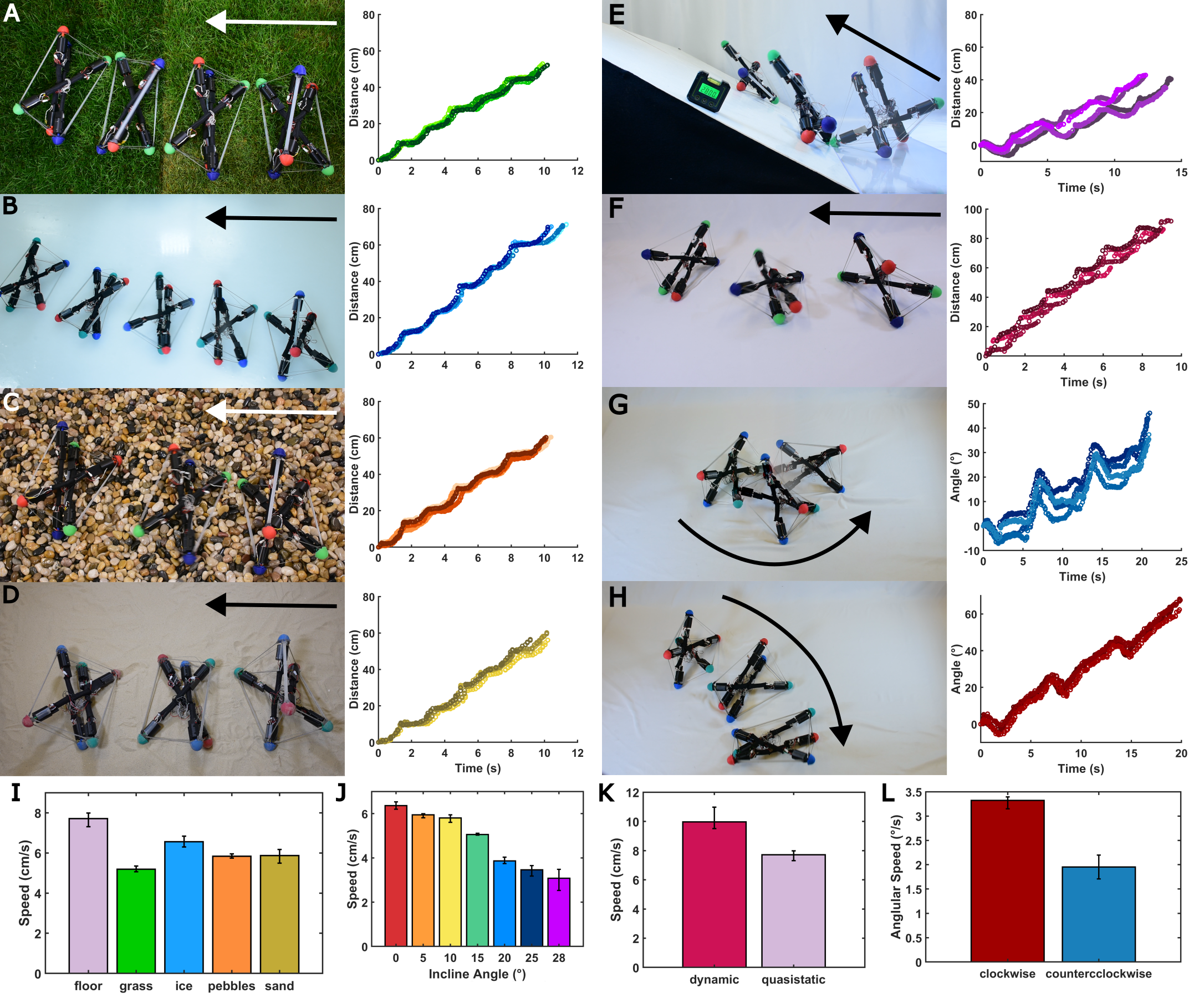}}
    \caption{\textbf{Locomotion Summary.}  The tensegrity robot can achieve locomotion across unstructured terrains, including (A) grass, (B) ice, (C) pebbles, and (D) sand.  It can also climb (E) inclines as steep as 28\degree.  Its top speed is achieved with its (F) dynamic rolling gait.  It can also turn (G) counterclockwise and (H) clockwise.  Each locomotion experiment is shown as a stitched photograph and a corresponding plot.  Arrows indicate the locomotion direction.  Each plot shows four trials of three cycles of the corresponding gait.  The locomotion speed is compared (I) across terrains, (J) as a function of incline angle (see Fig. S6), (K) for different rolling gaits, and (L) for different turning gaits (see Fig. S9).  The bar plots show the average speed over four trials while error bars represent the maximum and minimum speeds.
    % \textbf{Tensegrity locomotion on unstructured terrain and inclined surfaces.} Tensegrity locomotion trials and corresponding plots of the center of mass locations on various terrains, including (A) grass, (B) ice, (C) pebbles, and (D) sand.  Plots show four trials on the same terrain. (E) A comparison of the locomotion speed of the tensegrity robot on different terrains.
    % (F) A locomotion trial and corresponding plot of a tensegrity robot climbing a 28\degree incline.  The plot shows the center of mass location for four incline locomotion trials.  This experiment represents the steepest incline climbed by a tensegrity robot~\cite{chen2017inclined}. (G) A comparison of the locomotion speed of the tensegrity robot at different incline angles.  Further data from the incline experiments can be found in SUPPLEMENTARY MATERIALS.
    }
    \label{fig:locomotion}
\end{figure}

% MAKE AN ARGUMENT ABOUT HOW MOST WORK IS IN CONROLLED LAB SETTINGS

% reliable closed-loop control allows for robust locomotion over unstructured terrain and up inclines
%%% quasi-static rolling gait
%%% significance of 28 degree incline

The robot's locomotion was characterized across unstructured terrain,
% (Fig.~\ref{fig:locomotion}A-D and Fig.~\ref{fig:locomotion}I)
up inclines and down declines,
% (Fig.~\ref{fig:locomotion}E, Fig.~\ref{fig:locomotion}J, and Fig.~\ref{fig:inclines})
with different gaits for rolling
% (Fig.~\ref{fig:locomotion}F and Fig.~\ref{fig:locomotion}K)
and turning,
% (Fig.~\ref{fig:locomotion}G-H, Fig.~\ref{fig:locomotion}L, and Fig.~\ref{fig:turning})
and with shape morphing for varying the height of the robot. %body size.
% (Fig.~\ref{fig:shape_morphing})
The robot's quasistatic rolling gate is shown across different types of unstructured terrain, including grass, ice, pebbles, and sand in Fig.~\ref{fig:locomotion}A-D.  The corresponding plots show the robot's center of mass (CoM), which we define as the centroid of its six end caps, as a function of time over four trials.  The robot's speed across terrains is compared in Fig.~\ref{fig:locomotion}I.  Since this gait relies on gravity for rolling, the robot has comparable speeds across the different terrains, with grass being the most cumbersome.

The quasistatic rolling gait is a two-step, gravity-based rolling gait.  In the first step, the robot contracts two cables to transition onto the base triangle closer to the direction of rolling.  In the second step, the robot contracts three cables to deform its body, contracting its base triangle and moving its center of mass, in order to roll onto the next face.  Repeating this process achieves punctuated rolling.  For more details, see SM section M3.

% Enabled by the robot's reliable, closed-loop control, this gait can also be used to climb inclines as steep as 28\degree (Fig.~\ref{fig:locomotion}E), the steepest climbed by a tensegrity robot, as well as descend declines (movie S5).  The robot's speed across various incline angles is compared in Fig.~\ref{fig:locomotion}J wih more experiments shown in Fig.~\ref{fig:inclines}.

% Enabled by the robot's reliable, closed-loop control and shape morphing, t
The quasistatic rolling gait can also be used to climb inclines as steep as 28\degree 
 (Fig.~\ref{fig:locomotion}), the steepest climbed by a tensegrity robot.
% With feedback control, the robot can morph to increase its body size,
Feedback control enables the robot to morph to increase its height, %body size,
allowing the robot to ascend inclines effectively by taking larger steps with each cycle of the rolling gait.
% To ascend inclines effectively, the robot can morph to increase its body size and take larger steps with each cycle of the rolling gait. 
The robot's speed across various incline angles is compared in Fig.~\ref{fig:locomotion}J, and more incline experiments are shown in Fig.~\ref{fig:inclines}.
For controlled descent of declines (movie S5), the robot morphs to decrease its height and lower its CoM, preventing it from falling down the decline uncontrolled.  We characterize the robot's shape morphing ability in Fig.~\ref{fig:shape_morphing}, which shows the robot executing its quasistatic rolling gait with different heights.  This morphing ability enables the robot to navigate under obstacles (Fig.~\ref{fig:trajectory}D) by decreasing its height.

By skipping the transition step in the quasistatic rolling gait, the tensegrity robot exhibits dynamic rolling (Fig.~\ref{fig:locomotion}F) and achieves its top speed of 11~cm/s or 18 bar lengths per minute (BL/min).  The dynamic rolling gait is compared to the quasistatic rolling gait in Fig.~\ref{fig:locomotion}K.  Additional gaits for turning counterclockwise (Fig.~\ref{fig:locomotion}G) and clockwise (Fig.~\ref{fig:locomotion}H) were discovered in simulation~\cite{wang2022real2sim2real}, and they are compared in Fig.~\ref{fig:locomotion}L.  The asymmetry in the turning gaits is due to the robot's chiral topology.  The robot can also achieve a gradual turning gait by modulating its height to be larger on one side than the other.  More turning gaits are discussed in SM section S3 and shown in Fig.~\ref{fig:turning}.  Overall, this tensegrity robot has diverse locomotion capabilities; the gaits and their variations discussed in this section are the robot's motion primitives, the building blocks for navigation and control tasks.

% Feedback control enables the robot to modify not just its gait, but also its body size when executing a given gait.  Fig.~\ref{fig:shape_morphing} shows the robot executing its quasistatic rolling gait at different body sizes.  This morphing capability allows the robot to shrink in order to navigate under obstacles.
% Moreover, the robot can achieve a gradual turning gait (SM section S3) by modulating its size to be larger on one side than on the other.  Overall, this tensegrity robot has diverse locomotion capabilities; the gaits and variations thereof discussed in this section are the robot's motion primitives, the building blocks for navigation and control tasks.

% Reliable closed-loop control allows for robust locomotion over challenging terrain.  Fig. 3 shows the locomotion performance of the tensegrity robot over unstructured terrain, including grass, ice, pebbles, and sand.  The center of mass of the robot is tracked using our pose-tracking algorithm~\cite{lu20226n} for four trials on each terrain, demonstrating repeatable and reliable locomotion.  For these trials, we use a punctuated rolling gait.  This gravity-based gait exhibits comparable speeds on the different substrates.

% The robot can climb an incline of 28\degree (Fig. 1B), the steepest incline climbed by any tensegrity robot~\cite{chen2017inclined}.  The same punctuated rolling gait is used to climb inclines with varying slopes, and again we track the robot's center of mass.  The robot's velocity is greater on shallower inclines (see supplementary materials).

\section*{Autonomous Control}

% figure four
% model predictive control results for trajectory following

\begin{figure}
    \centering
    \makebox[\textwidth][c]{\includegraphics[width=\linewidth]{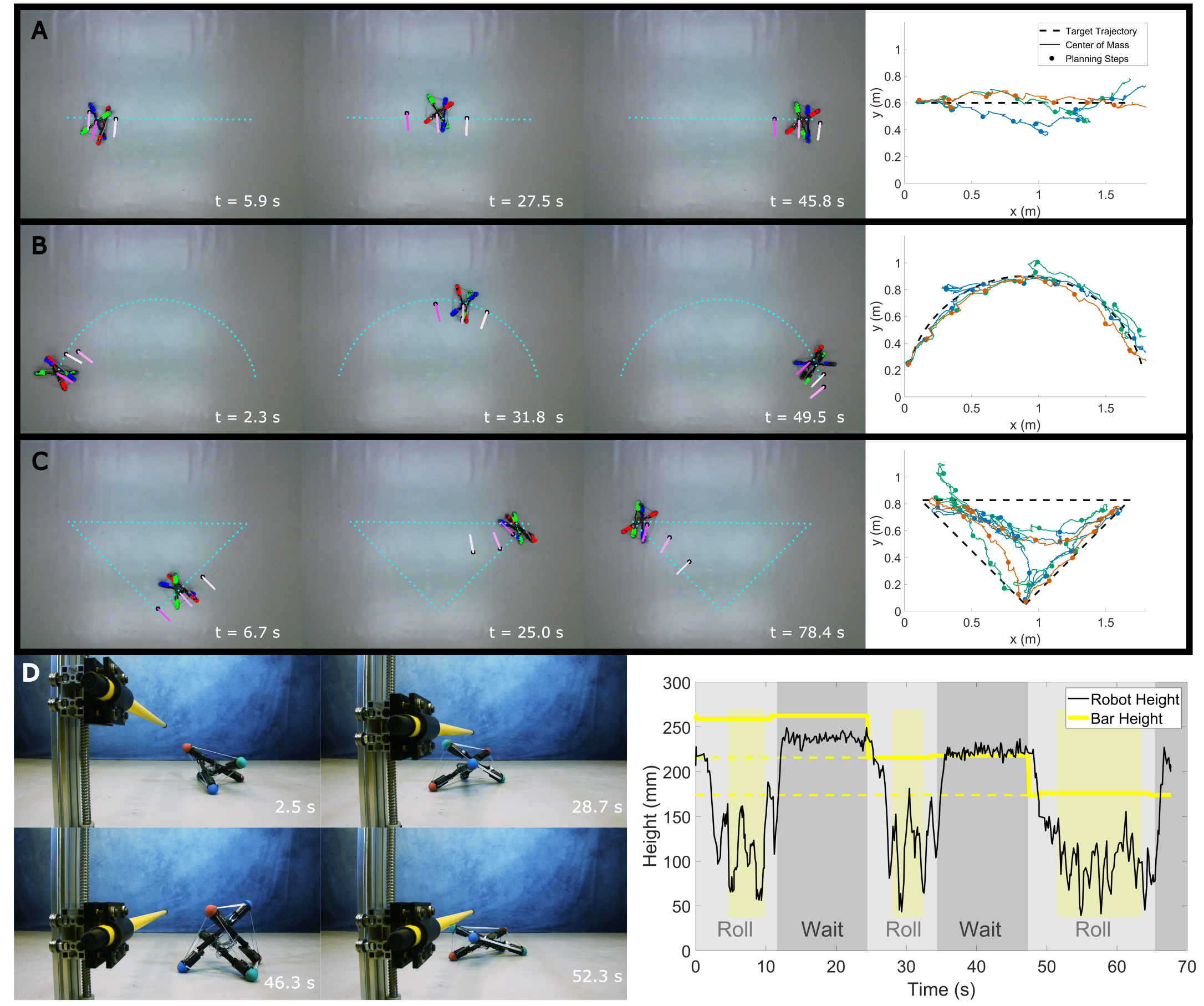}}
    \caption{\textbf{Autonomous control.}
    % (A) The tensegrity robot plans a sequence of actions, shown with black markers for the predicted center of mass and magenta vectors for the predicted principal axis, that enable it to best follow the prescribed trajectory plotted in cyan.  (A) A trial where the tensegrity robot tracks a straight line trajectory. (B) A plot of four trials of the tensegrity robot tracking a straight line trajectory.  The plot shows the target trajectory and the robot's center of mass, with the robot's center of mass at planning steps shown with circular markers. (C) A semicircle trajectory experiment. (D) Five trials of tracking a semicircle trajectory. (E) Tracking a triangular trajectory. (F) Four trials of tracking a triangular trajectory.
    (A)-(C) The tensegrity robot autonomously follows a provided trajectory via a model-based controller.  The robot plans a sequence of actions---shown with black markers for the predicted center of mass (CoM) and magenta vectors for the predicted principal axis---that enable it to best follow the trajectory plotted in cyan.  The darkest magenta vector corresponds to the robot's measured starting state and each lighter vector to the successive states predicted for following the planned action sequence.  Each corresponding plot shows the target trajectory and robot's CoM for three trials.  Circular markers indicate the CoM at planning steps.  The tracked trajectories include (A) a straight line, (B) a circular arc, and (C) a right triangle.  (D) The tensegrity robot autonomously plays limbo, sensing the height of the yellow bar with an off-board camera and adapting its body height to roll under it.  The corresponding plot shows the height of the robot compared to the bar height during the limbo demonstration.  The gray shaded regions show when the robot is executing its rolling gait and when it is waiting for the bar to lower while the yellow shaded regions show when part of the robot is directly under the bar.}
    \label{fig:trajectory}
\end{figure}

% Autonomous navigation has yet to be shown in tensegrity robots~\cite{shah2021tensegrity} as most tensegrity robots only demonstrate open-loop trajectories.  Combining the low-level PID controller with our pose tracking algorithm~\cite{lu20226n} and a simulation based on differentiable physics we developed in previous work~\cite{wang2022real2sim2real}, we construct a tiered autonomous controller that enables the robot to reliably follow provided trajectories.

Tensegrity robots in the reported literature demonstrate open-loop trajectories.  Here, we introduce a model-based, autonomous controller that enables closed-loop trajectory following (Fig.~\ref{fig:trajectory}A-C).  Tensegrity robots are difficult to model accurately due to their complex, coupled dynamics and many degrees of freedom~\cite{shah2021tensegrity}.  In previous work, we presented a Real2Sim2Real framework for modeling tensegrity robots based on a differentiable physics engine~\cite{wang2022real2sim2real}.  In this work, this simulator was used to model each of the tensegrity robot's motion primitives (the distinct gaits and their variations discussed above) to build a database of state-action transitions.  The robot's position and orientation are tracked in real time with an overhead RGB-D camera using a tensegrity pose-tracking algorithm adapted from our previous work~\cite{lu20226n}.  Given the robot's state, the controller performs a two-ply exhaustive search to determine the sequence of actions that minimizes the cost function of the trajectory following task~\cite{nagabandi2018learning,huang2022design}.  After executing the first action in the sequence and subsequently measuring the robot's state, the controller re-plans a new sequence of two actions.  More details are available in SM section M4.

The results of the trajectory following experiments are shown for a straight line, a circular arc, and a right triangle trajectory in Fig.~\ref{fig:trajectory}A-C.  The images show the robot with the pose tracking, the trajectory, and the planned action sequence superimposed, with successively lighter colors corresponding to predictions further in the future.  The corresponding plots show the robot's CoM over three trials for each trajectory.  Larger markers represent planning steps.  The triangular trajectory is segmented into three straight line trajectories, and the robot proceeds to the next trajectory segment in the sequence when its CoM falls within a given radius of the end of the current segment.  At that moment, the controller determines if the robot will follow the next segment in the same direction or if it will reverse its direction and modify its gait via symmetry reduction (SM sections M4 and S6).  The error (Fig.~\ref{fig:trajectory_error}) is higher for the latter two segments of the triangular trajectory because the robot starts those segments where it ended the previous ones rather than with the perfect starting position and orientation.

Fig.~\ref{fig:trajectory}D and movie S7 show another demonstration of autonomy: a game of limbo where the robot repeatedly navigates under an obstacle of decreasing height.  The overhead camera tracks the robot's pose and the height of the bar in real time.  After the robot reaches the other side, the bar lowers, and the robot reverses its gait (SM section S6) and restricts its range to keep its body size below the bar height.  The plot in Fig.~\ref{fig:trajectory}D shows the robot height and bar height during the trial.  The yellow shaded regions indicate when the robot is under the bar while the gray shaded regions show whether the robot is executing its rolling gait or waiting for the bar to lower.

The control tasks demonstrated in this work are the building blocks for future tensegrity robots that can navigate unstructured environments.  Because human teleoperators may not always have open lines of communication, these robots must be capable of executing their most fundamental tasks autonomously, like following the last provided path and avoiding obstacles.  Furthermore, the impact resistance of tensegrity robots can enable more efficient path planning~\cite{shah2021tensegrity}.  Instead of navigating around cliffs and drops, our robot can roll off and survive the fall (Figs.~\ref{fig:splash} and~\ref{fig:limbo_cliff}).

\section*{Impact Resistance}

% figure five
% limbo and cliff dive frames and diagrams

\begin{figure}
    \centering
    \makebox[\textwidth][c]{\includegraphics[width=\linewidth]{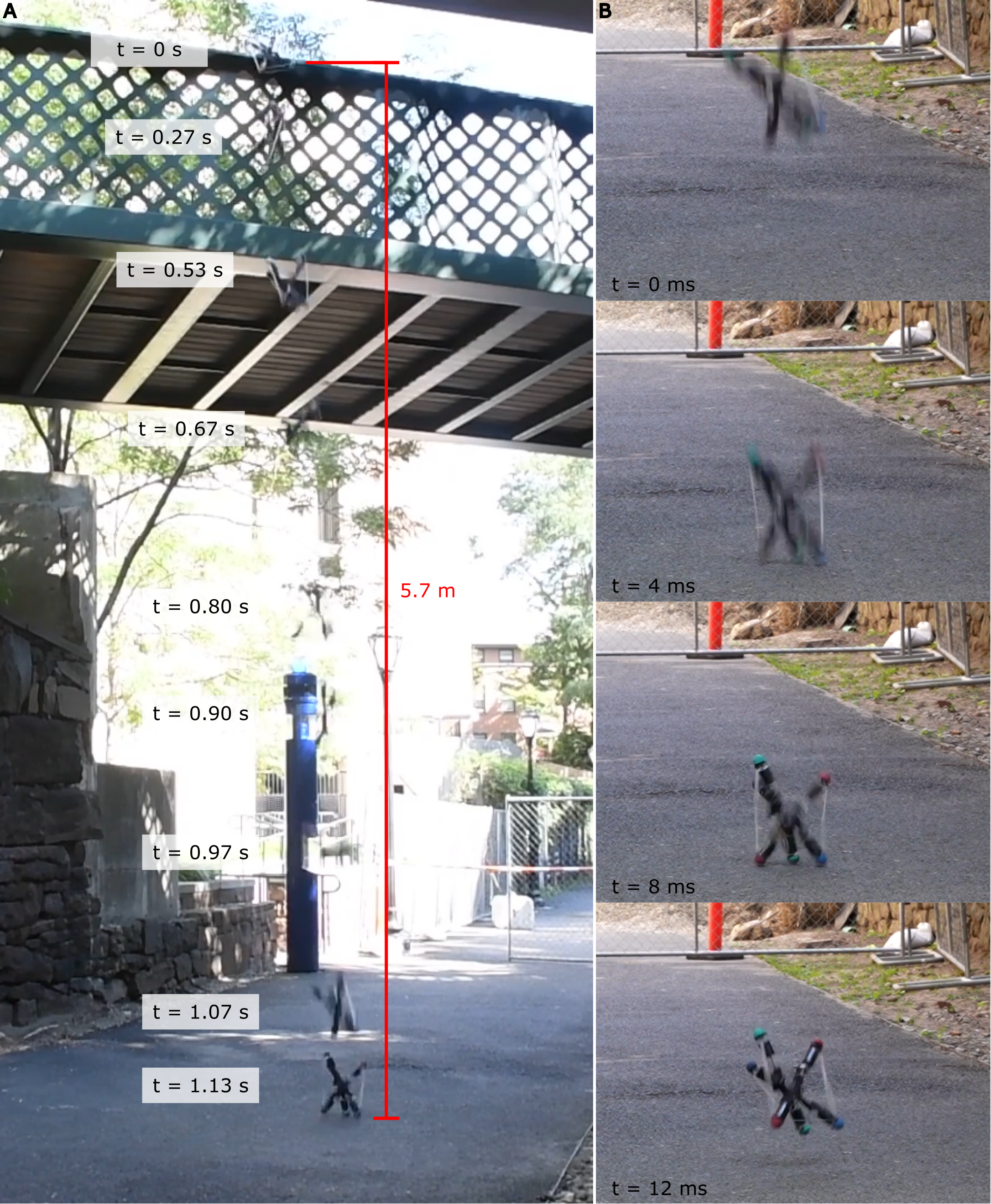}}
    \caption{\textbf{Impact resistance.}
    % (A) The tensegrity robot senses the height of the obstacle and adapts its body height to roll under the bar in a game of limbo. (B) A plot of the robot and bar heights during the limbo trial.  The robot's body shrinks to successfully pass under the bar during the time it is rolling. (C)
    The robot survives a 5.7~m drop after rolling off a bridge.  Movie S8 shows the robot continuing its locomotion after surviving the fall.}
    \label{fig:limbo_cliff}
\end{figure}

% Autonomous tensegrity robots can leverage their extreme deformations and impact resistance to navigate obstacles that would be challenging or intractable for other mobile robots.  We demonstrate this autonomous capability with our robot by playing limbo, a game where the robot must go under a bar with ever-decreasing height, and rolling off of a cliff.  

% autonomous shape morphing to avoid obstacles
%%% bar height detection
%%% gait modification

% To play limbo, the robot tracks its position and the current height of the bar using the pose-tracking algorithm~\cite{lu20226n} and an RGB-D camera.  Then, the robot executes its punctuated rolling gait, limiting its range of motion depending on the current bar height, until it reaches the other side.  As the bar height lowers, the robot's range of motion decreases to keep its height low enough to successfully navigate under the bar (Fig. 5A-B).

% The tensegrity robot is also notable for its impact resistance.  The robot survives a drop from a bridge (Fig. 1C) after rolling off the edge.  Fig 5C shows how the tensegrity robot absorbs the energy from the impact through extreme deformations of the structure, enabling the robot to survive the crash (see materials and methods).

% \section*{Conclusion}

% In conclusion, we present a high-speed, multi-gait, multi-terrain, autonomous tensegrity robot that is capable of surviving harsh impacts.  This work paves the way for future mobile robots that can autonomously navigate perilous, remote environments.

The ability to survive harsh impacts is one of the most noteworthy advantages of tensegrity robots, yet only a few researchers demonstrate their robots enduring drop tests~\cite{vespignani2018design,rieffel2018adaptive,zappetti2022dual,spiegel2023shape}.  In addition to the fall from the 2~m cliff shown in Fig.~\ref{fig:splash}, movie S8 and Fig.~\ref{fig:limbo_cliff} show our tensegrity robot rolling off a 5.7~m bridge, surviving the subsequent shear drop---the highest drop by a tensegrity robot in the reported literature---and then continuing its rolling locomotion (SM section M6).  Movie S11 shows the robot surviving a fall from a larger cliff in an authentic field environment, though it is not a shear drop as the robot is buffeted by vegetation and the cliff face on the way down (SM section M6).

The bulk compliance afforded by the robot's tensegrity structure allows it to absorb the energy of the impact through deformation.  As the external load is applied, the struts reorient such that all are loaded in pure axial compression while the tendons are loaded in pure axial tension.  The ability to absorb large impacts from falls could one day enable tensegrity robots to explore the bottoms of craters on planetary surfaces or to be quickly deployed to disaster zones by dropping them from airplanes.

We envision a future where robots are versatile, robust, and autonomous so that they may navigate and explore environments that are out of reach for today's state-of-the-art robots.  Part of the solution is introducing new hardware platforms with enhanced capabilities.  Inspired by art and architecture, we have created a 3-bar tensegrity robot to push boundaries.  This tensegrity robot has reliable, highly stretchable sensors and for state estimation and autonomous control.  The robot is versatile, navigating various terrains and obstacles with several locomotion gaits, and its robustness is demonstrated by subjecting it to harsh impacts.  We expect this tensegrity robot to serve as a platform and a benchmark for further studies as we work toward a future with capable, compliant robots.

\clearpage

\bibliography{main}

\bibliographystyle{Science}

\section*{Acknowledgments}
W. R. J., X. H., J. W. B., and R. K. B. were supported by the National Science Foundation (NSF) under grant no. IIS-1955225.  S. L., K. W., and K. B. were supported by the NSF under award IIS-1956027.  The authors would like to thank Frank Butler and the staff at the Ingalls Rink for the generous use of their ice for the multi-terrain experiments.  Additionally, we would like to thank John Campbell, Sean O'Grady, Stephanie Woodman, Luis Ramirez, Monica Li, and Caitlin Le for their assistance with the large cliff experiments in East Rock Park.
% Include acknowledgments of funding, any patents pending, where raw data for the paper are deposited, etc.

%Here you should list the contents of your Supplementary Materials -- below is an example. 
%You should include a list of Supplementary figures, Tables, and any references that appear only in the SM. 
%Note that the reference numbering continues from the main text to the SM.
% In the example below, Refs. 4-10 were cited only in the SM.     
\section*{Supplementary materials}
Materials and Methods\\
Supplementary Text\\
Figs.~\ref{fig:endcaps} to~\ref{fig:symmetry}\\
Tables~\ref{tab:terrain_stats} to~\ref{tab:turn_stats}\\
Movies S1 to S11\\

% For your review copy (i.e., the file you initially send in for
% evaluation), you can use the {figure} environment and the
% \includegraphics command to stream your figures into the text, placing
% all figures at the end.  For the final, revised manuscript for
% acceptance and production, however, PostScript or other graphics
% should not be streamed into your compliled file.  Instead, set
% captions as simple paragraphs (with a \noindent tag), setting them
% off from the rest of the text with a \clearpage as shown  below, and
% submit figures as separate files according to the Art Department's
% instructions.

\clearpage

\section*{Materials and Methods}
\renewcommand{\thesubsection}{M\arabic{subsection}}

\renewcommand{\thefigure}{S\arabic{figure}}
\setcounter{figure}{0}

\renewcommand{\thetable}{S\arabic{table}}
\setcounter{table}{0}

\subsection{Robot Design}
% diagram/overview

\subsubsection{Mechanical Components}

\begin{figure}
    \centering
    \makebox[\textwidth][c]{\includegraphics[width=4.5in]{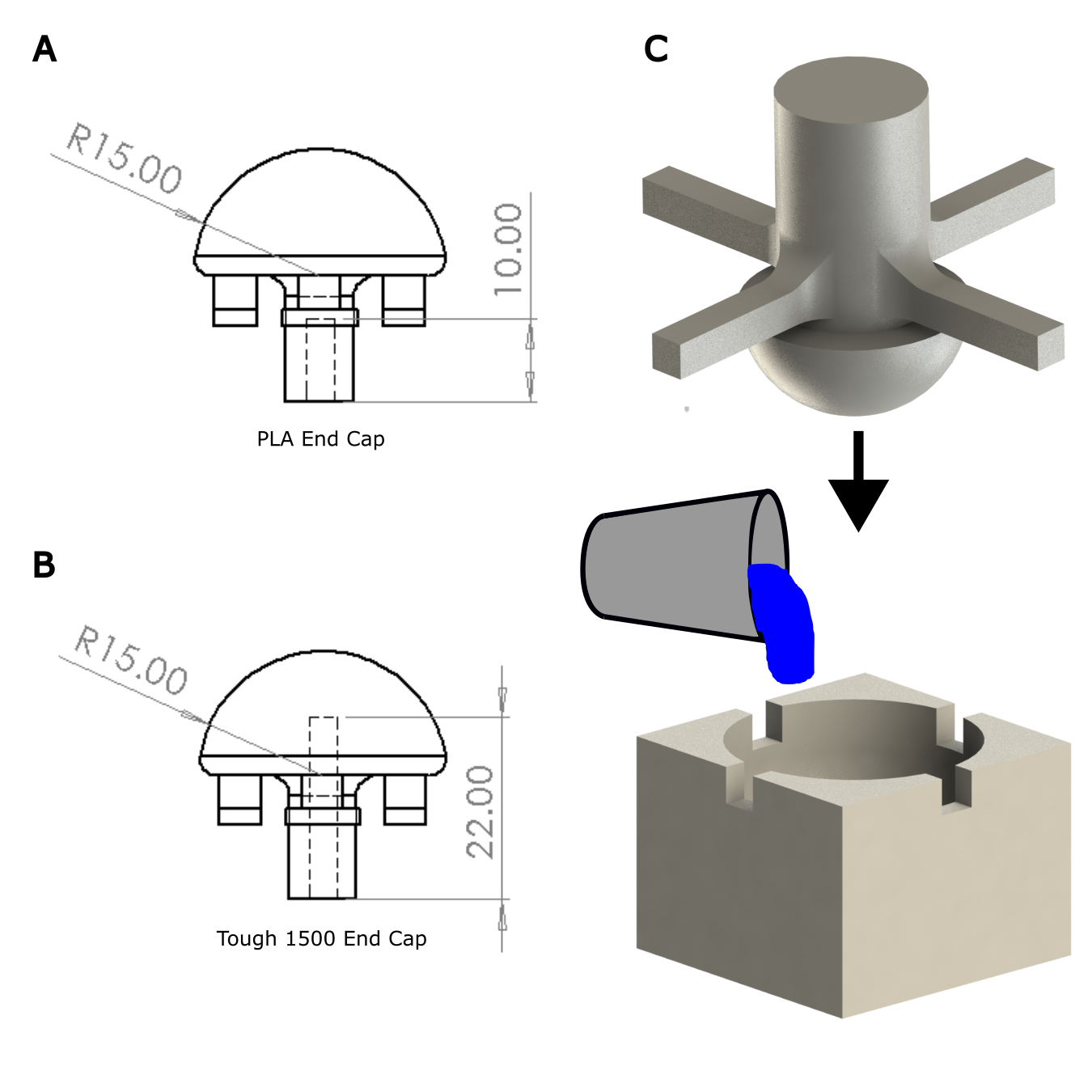}}
    \caption{\textbf{Mechanical design of end caps.} (A) Schematic of PLA end cap. (B) Schematic of Tough 1500 end cap used for cliff and bridge experiments. (C) A schematic showing the two-part molding process for the silicone covers using DragonSkin 30 (Smooth-On).  The files for 3D printing the end caps and the mold pieces are available in the supplementary data.}
    \label{fig:endcaps}
\end{figure}

The robot comprises three 1/8'' carbon fiber rods (2153T16; McMaster-Carr) with mounted electronics, 3D printed end caps, and cables and silicone tendons that connect the end caps to form the tensegrity structure (Fig. 1D).  For the majority of the experiments, the end caps are 3D printed from polylactic acid (PLA), and their dimensions are shown in Fig.~\ref{fig:endcaps}A.  The rods are cut to length so the overall length of one bar is 360~mm from end cap to end cap. For the impact resistance experiments (section M6), the end caps are 3D printed from Tough 1500 resin (FormLabs) with the dimensions shown in Fig.~\ref{fig:endcaps}B.  Note that the rods in the impact resistance experiments are 12~mm shorter on each end due to the change in end cap dimensions.  Each rod has two 3D printed clamping mounts, one on either end, that each houses a 12~V High-Power Carbon Brush Micro Metal Gearmotor with a 150:1 gear ratio (3042; Pololu) and a 220~mAh 1S LiPo battery.  The 3D printed mounts have heat-set inserts (94459A250; McMaster-Carr) that add threads to the plastic parts such that they can be clamped onto the rod and held in place by tightening two 4-40 screws.

Each motor has on its shaft a 3D printed winch with a 7~mm diameter that extends and contracts fishing line (Power Pro) that connects the motor to an end cap on another bar.  Each 3D printed end cap has four hooks: one where the fishing line from a motor attaches and three others where sensor tendons, described below, attach.  Each end cap also has a silicone cover to enhance the friction and dampen the footfalls of the robot's end caps. The covers are made by pouring a silicone elastomer (DragonSkin 20; Smooth-On) in a 3D printed mold (Fig.~\ref{fig:endcaps}C.  The end caps and their corresponding covers are colored red, green, and blue to enable pose tracking of the robot, described in our previous work~\cite{lu20226n} and summarized in Supplementary Materials (SM) section S4.

The six 1S batteries distributed across the rods are wired together 3S2P for a nominal 11.1~V to drive the 12~V motors.  The electronics, described below, are housed in 3D printed cases with 1/8'' holes that allow them to slide onto the rods.  Wires are wrapped in protective sleeving.  All 3D printed parts are made from polylactic acid (PLA), and the files for printing them are provided in the supplementary data.

\subsubsection{Electronics}

The robot's on-board electronics comprise a custom control board containing a microcontroller with a Bluetooth module (BMD-350), a voltage regulator (AP2112K), and three motor drivers (TB6612FNG); a capacitive sensing breakout board (MPR121; Adafruit); an I2C multiplexer breakout board (TCA9548A); and a breakout board for a 9-axis IMU with an on-board digital motion processor (ICM-20948; Pimoroni).  The design for the control board is available for download in the supplementary data.  A schematic is shown in Fig.~\ref{fig:electronics}.  The electronics are distributed among the three rods and connected with wires.  The red rod hosts the control board, the green rod the MPR121, and the blue rod the mutliplexer.  Each rod contains one IMU and two batteries.  All the peripherals are connected to the control board via I2C (the IMUs are connectiated via the multiplexer).  The motors and batteries connect directly to the control board while the sensor tendons are connected to the MPR121.

\begin{figure}
    \centering
    \makebox[\textwidth][c]{\includegraphics[width=7.24in]{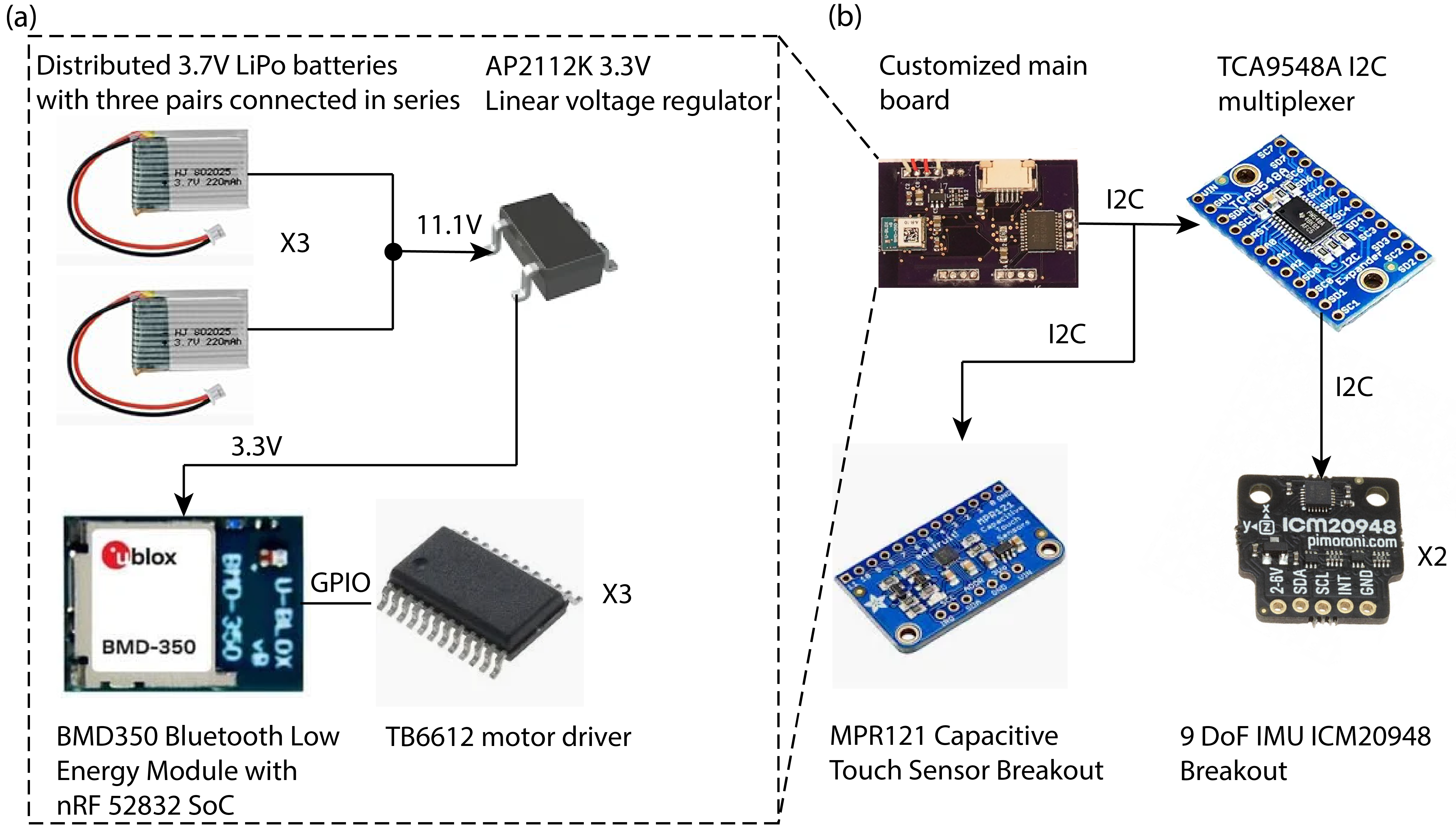}}
    \caption{\textbf{Electrical design.}
    % (A) The schematic of the design of the control board, the custom PCB with the microcontroller and motor drivers. (B)
    A schematic of how the electronic components on the robot are connected. This circuitry is distributed among the three bars of the robot.  Each rod contains one IMU, but only two of the three are used for state estimation. The design files for the customized main board are available in the supplementary data.}
    \label{fig:electronics}
\end{figure}

\subsubsection{Capacitive Sensor Tendons}

We have previously introduced capacitive sensor tendons~\cite{johnson2022sensor}, capacitive strain sensors that double as the elastic tendons of a tensegrity robot.  Sensor tendons comprise two layers of a liquid metal paste~\cite{wang2019printed,kong2020oxide} encapsulated by a silicone elastomer (Smooth-On) to form a stretchable parallel-plate capacitor.  As the tendon is stretched, the capacitance increases linearly~\cite{shintake2018ultrastretchable}, so we can know the length of the tendon by measuring its capacitance with the MPR121.

There are two types of sensor tendons: thinner tendons run parallel to the cables driven by the motors, and thicker tendons run parallel to the long axis of the robot.  The thinner tendons need to accommodate high strains so that they do not limit the extreme deformations necessary for tensegrity locomotion while the thicker tendons need to exert high forces since they act like passive springs responsible for restoring the robot to its original shape after deformation.  The dimensions of the two types of sensor tendons are shown in Fig.~\ref{fig:sensors}, and the files necessary for fabricating them are available in the supplementary data.  Further characterization and manufacturing details are available in the original paper~\cite{johnson2022sensor}.  After the sensors are manufactured, their wire leads are extended by 140~mm with ultra flexible 29 AWG wire (9564T2; McMaster-Carr) and crimped with Dupont connectors.

% calibration procedure
Each of the nine sensor tendons is calibrated by stretching it to five known lengths, measuring its corresponding capacitance at each length (with the MPR121), and fitting a linear mapping between capacitance and length for that sensor.  An example calibration is shown in Fig. S3.  During state reconstruction and locomotion, described below, the capacitance of each sensor is constantly measured, and the linear mapping obtained from the calibration is used to calculate each tendon's length for shape estimation and feedback control.  The sensors need to be recalibrated at least once per day because wires shifting positions during locomotion trials or repairs can change the baseline capacitance measured by the MPR121 for each sensor.

\begin{figure}
    \centering
    \makebox[\textwidth][c]{\includegraphics[width=5.5in]{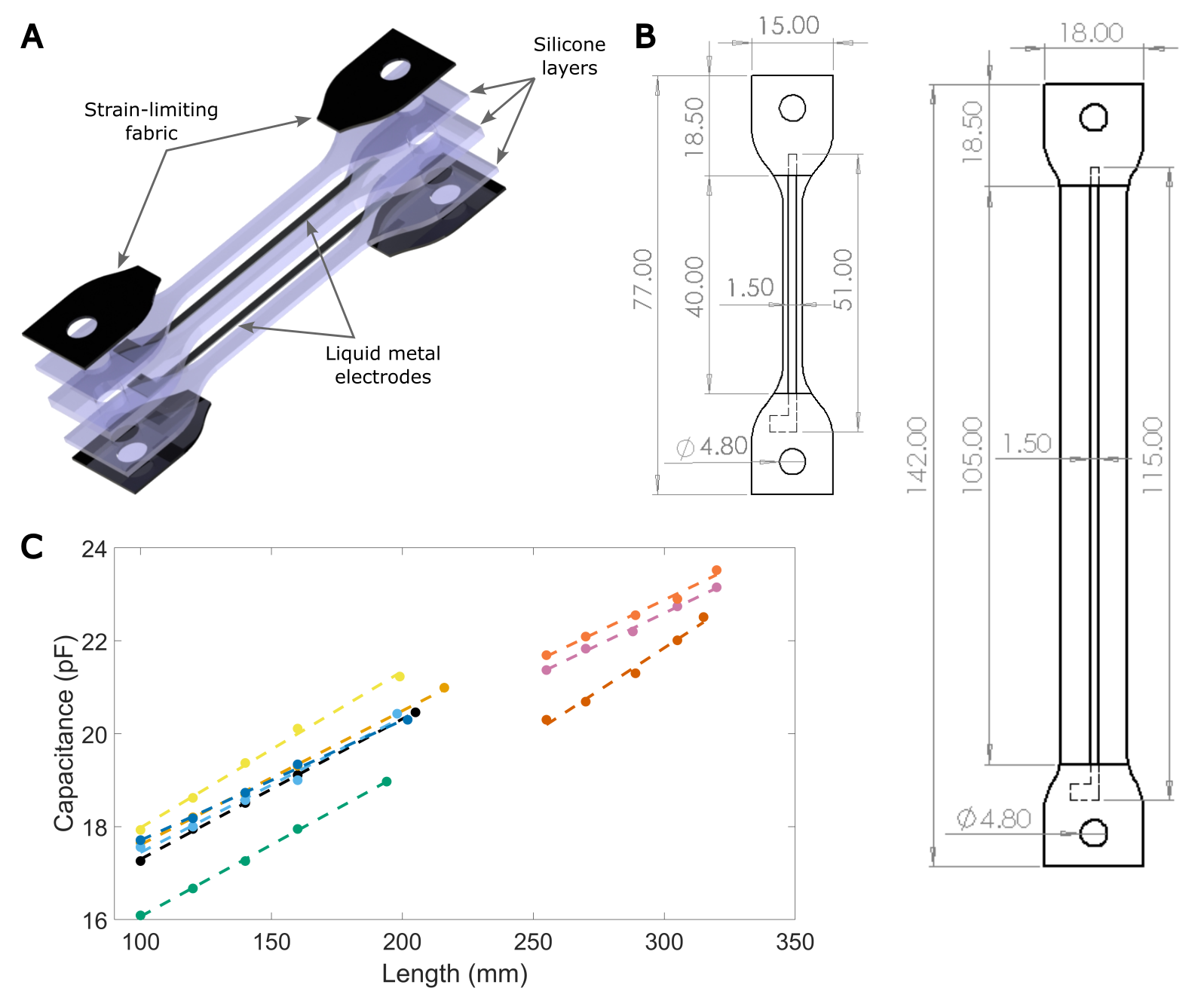}}
    \caption{\textbf{Sensor tendons.} (A) An exploded view of the shorter type of sensor tendon showing how the liquid metal electrodes and strain-limiting fabric are encapsulated in the silicone.  (B) A schematic of the two types of sensor tendons.  The longer, thicker sensors double as passive springs that restore the robot's original shape after actuation.  The shorter, thinner sensors run in parallel with actuated cables and provide feedback for closed-loop control. (C) A representative example of sensor calibration.  Each tendon on the robot is stretched to five known lengths, and the corresponding sensor's capacitance is measured.  Capacitance varies linearly with length for each sensor, but each sensor has a different calibration curve.}
    \label{fig:sensors}
\end{figure}

\subsection{State Estimation}

\begin{figure}
    \centering
    \makebox[\textwidth][c]{\includegraphics[width=7.24in]{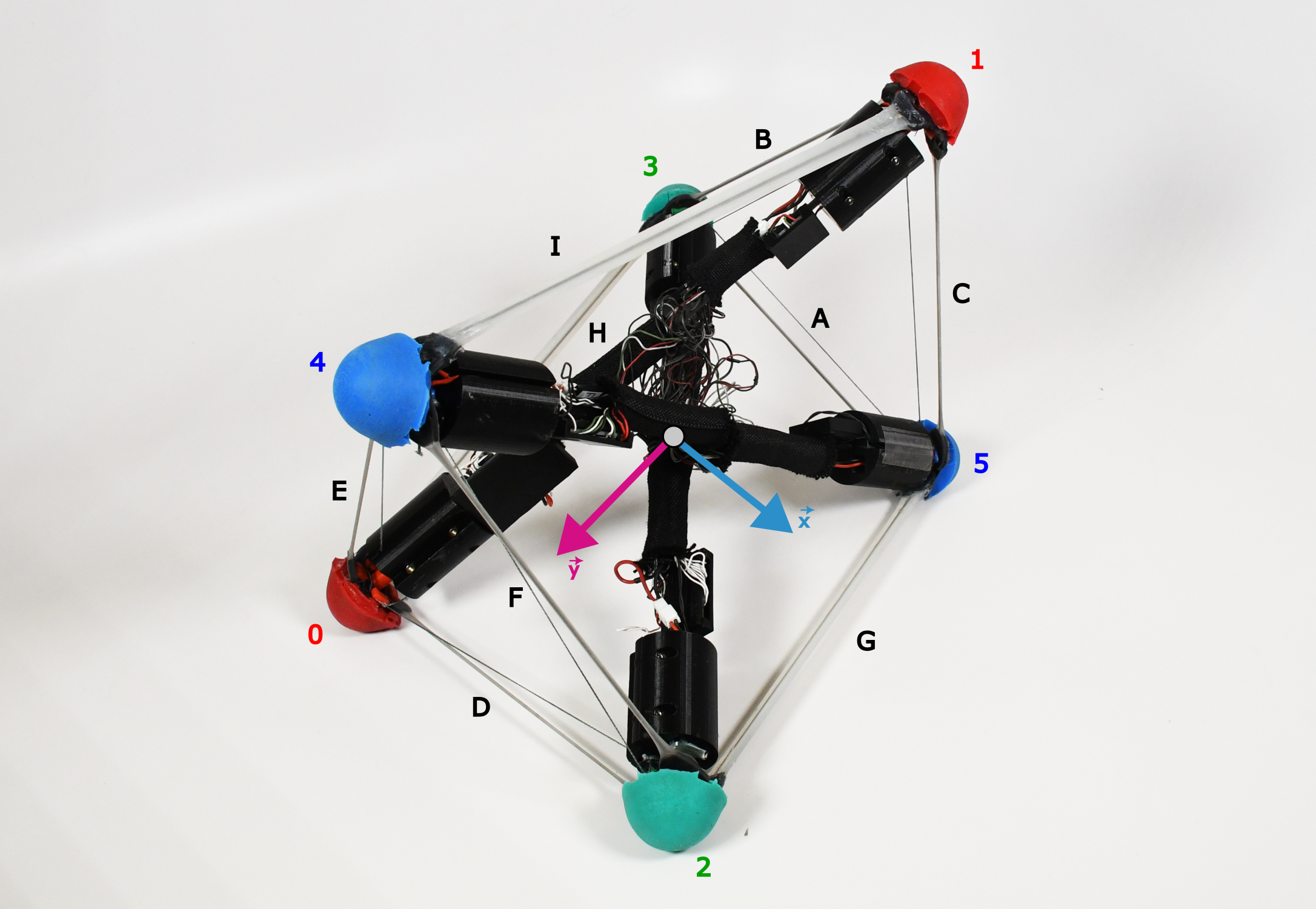}}
    \caption{\textbf{Robot labeling scheme.} A schematic showing the naming convention for the nodes and sensors that is used throughout this paper.  The robot is shown in its rest state, from which all its gaits are defined, meaning the actuated tendons are all 200~mm in length and nodes 0, 2, 3, and 5 are the closest to the ground.  For the state reconstruction experiments, the definition of the robot's body-frame x- and y-axes are shown.  The y-axis is parallel to the vector that points from the centroid of nodes 1, 3, and 5 to the centroid of nodes 0, 2, and 4.  The x-axis is parallel to the vector that points from node 0 to node 2.  The z-axis is defined based on x and y with a right-hand convention.  For the trajectory following experiments, the robot's principal axis is the 2D projection of the y-axis shown here.}
    \label{fig:labels}
\end{figure}

Advancing beyond what was demonstrated in previous literature~\cite{caluwaerts2016state,booth2020surface,li2021shape}, Tribar is capable of reconstructing its state during dynamic locomotion using only on-board sensors.  Here state is defined as the robot's shape plus its orientation relative to a global frame.  The robot's on-board sensors include the nine strain sensors that measure the length of the nine tensegrity tendons and three 9-axis IMUs.  Each IMU contains a 3-axis accelerometer, a 3-axis gyroscope, and a 3-axis magnetometer.  These three sensors along with the IMU's on-chip digital motion processor are used to measure the IMU's orientation relative to Earth's magnetic and gravitational fields.  Two such measurements from IMUs on different bars, estimating each bar's direction in a global frame, provide enough information for state estimation.

Our state estimation algorithm is a two-step optimization algorithm.  The first step is shape estimation where we minimize the squared error from the sensor tendon measurements in a constrained optimization given the known bar lengths, as in ~\cite{booth2020surface}.  Precisely, shape estimation minimizes the objective
\begin{center}
\begin{math}
    \frac{\sum^2_{i=0}\sum^2_{j=0,i\neq j}
    \left( \left|\left| N_{2i+1} - N_{2j+1} \right|\right| - d_{2i+1,2j+1} \right)^2 
    + \left(\left|\left| N_{2i} - N_{2j} \right|\right| - d_{2i,2j}\right)^2
    }{2}
    + \sum^2_{k=0}
    \left(\left|\left|N_k - N_{k+3} \right|\right| - d_{k,k+3}\right)^2
\end{math}
    \\
\begin{math}
    \textrm{s.t.} \left|\left| N_{2l} - N_{2l+1} \right|\right| - L = 0 \ \textrm{for} \ l = 0,1,2
\end{math}
\end{center}
where $N_m$ is the 3D position of node $m$, $d_{p,q}$ is the sensor measurement between nodes $p$ and $q$, and $L$ is the constant bar length.  The node labeling scheme is shown in Fig.~\ref{fig:labels}.  Nodes 0 and 1 are on the red bar, 2 and 3 on the green, and 4 and 5 on the blue.  The odd numbered nodes are all on one side of the robot while the even numbered nodes are on the opposite side.  The output of the shape estimation step is the 3D position vector of each node in an arbitrary frame whose origin, for convenience, is at the centroid of the six nodes.

The second step of the algorithm is orientation estimation where, given the shape of the nodes, we find the rotation that minimizes the squared error between the unit vectors of the estimated and measured directions of the two bars with IMUs relative to the global frame.  Precisely, we find the rotation $R$ that minimizes the objective
% \begin{center}
\begin{equation*}
% \operatorname*{argmin}_R
\sum^1_{l=0} \left( R \frac{N_{2l} - N_{2l+1}}{\left|\left| N_{2l} - N_{2l+1} \right|\right|} - \hat{a}_l \right)^2
\end{equation*}
% \end{center}
where $N_m$ is again the 3D position of node $m$ and $\hat{a}_l$ is the unit vector representing the orientation of bar $l$ as measured by the IMU.  Once $R$ is found through the optimization routine, we apply the rotation to the nodes and return the rotated coordinates.  These rotated coordinates represent the robot's state, defined as its shape and orientation relative to a global frame.  Note that our algorithm does not estimate the robot's translation.

\begin{figure}
    \centering
    \makebox[\textwidth][c]{\includegraphics[width=7.24in]{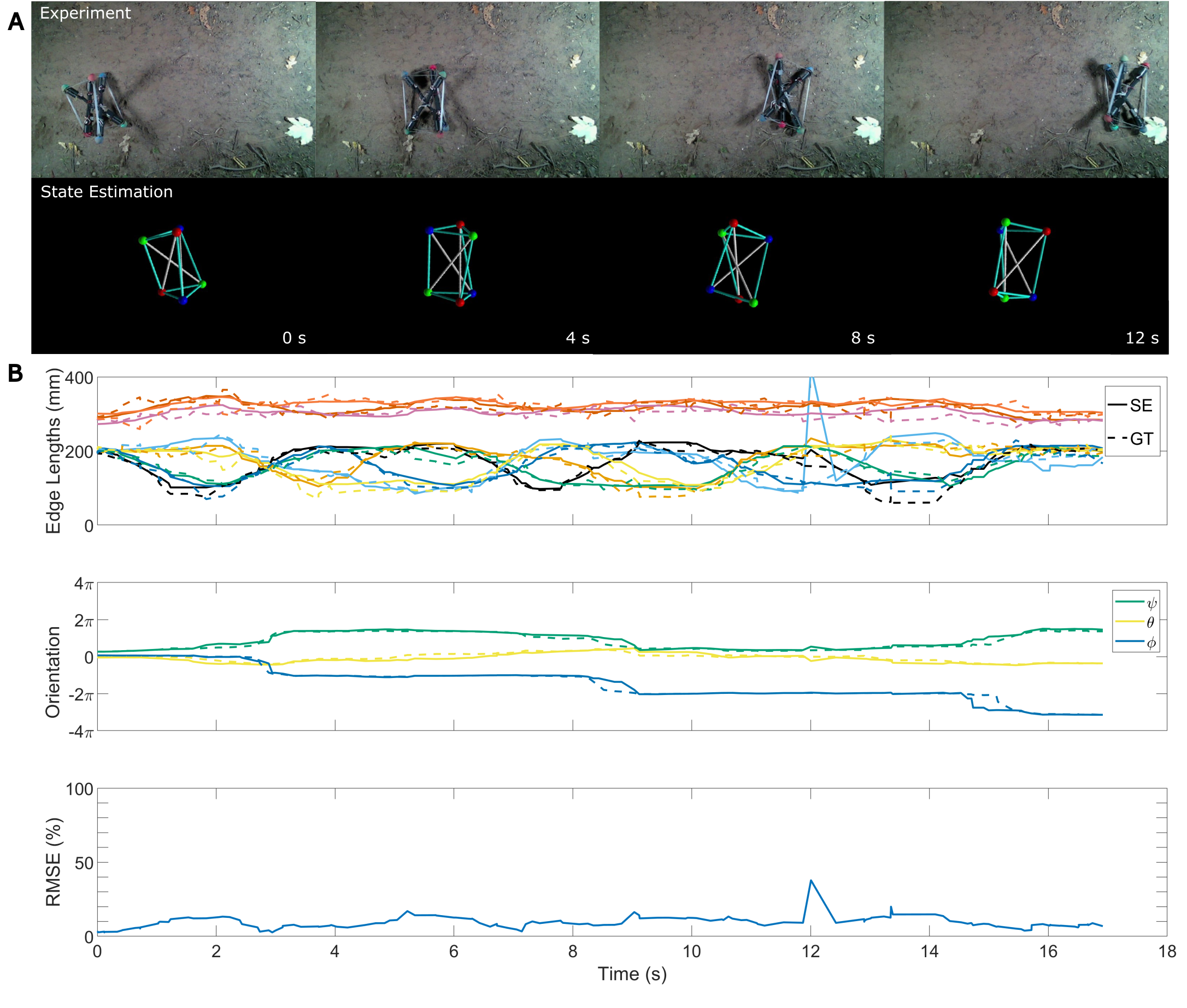}}
    \caption{\textbf{State estimation during forward locomotion.}
    % (A) Rendering of sensor reconstructions compared to video frames from the experiment. (B) Graph of state reconstruction.
    (A) A locomotion trial in a field setting with photographs
    and corresponding renderings of the robot’s estimated state from onboard sensors. (B) A plot of the same locomotion trial showing the results of state estimation (SE) from onboard sensors compared to ground truth (GT) from our vision-based pose tracking algorithm (55). The top subplot shows the nine tendon lengths, the middle subplot shows the ZYX Euler angles that describe the robot’s orientation relative to the global frame defined by Earth’s gravitational and magnetic fields, and the bottom subplot shows the root mean square error (RMSE) between the estimated and ground truth positions of the nodes relative to the bar length. The robot executes its quasistatic rolling gait.}
    \label{fig:state_reconstruction2}
\end{figure}

The robot's state reconstruction performance during dynamic locomotion trials in a field environment is shown in Fig.~\ref{fig:state_recon} and Fig.~\ref{fig:state_reconstruction2}.  A field environment was chosen to avoid magnetic interference from nearby buildings.  The magnetic interference disrupts the readings of the IMU's magnetometer and therefore the output of the orientation estimation.  The field experiments were conducted at dusk under artificial lighting from portable LED spotlights so that the sunlight did not interfere with the Intel RealSense L515 depth camera, which we used to track the robot's position for ground truth~\cite{lu20226n}.

\subsection{Locomotion}
% experimental setup (ROS, cameras, etc.)
The tensegrity robot achieves locomotion through extreme deformations that shift its center of mass.  Sensor tendons (section M1.3) provide real-time feedback for PID control (described below) so that each of the six actuators can achieve a target length within some small tolerance.  A set of six target lengths is a target shape, and a repeating sequence of target shapes defines a gait.  The robot has several different gaits (described below) for rolling and turning locomotion.

\subsubsection{PID Control}
Each gait is parameterized by a {\em minimum length} and a {\em range} for each actuator.  The {\em position} of each actuated tendon is the normalized length of the tendon where $0$ is the minimum length and $1$ is the minimum length plus the range.
\begin{equation*}
    \textrm{position} = \frac{\textrm{length} - \textrm{mininum length}}{\textrm{minimum length} + \textrm{range}}
\end{equation*}
For a given target shape, each of the six actuated tendons will have a target length between $0$ and $1$.  The error between the current position and the target is given by $e = \textrm{position} - \textrm{target}$, and the command to each motor is calculated via PID control.
\begin{equation*}
    u_T = K_P e_T + K_I \sum_{t=0}^T e_t + K_D (e_T - e_{T-1})
\end{equation*}
At each time step, the calculated command $u_T$ is then multiplied by another gait parameter, {\em max speed}, and the product is bounded by $\left[-\textrm{max speed}, \textrm{max speed}\right]$ where a result of $99$ (or~$-99$) corresponds to the maximum duty cycle sent to the motor.

\subsubsection{Gaits}

The PID controller described above ensures all six actuated tendons are set to their target length (a normalized position between $0$ and $1$), within some small tolerance, given as a percent of the range.  Together, these six target lengths define a target shape, which is one step in a gait, a repeating sequence of target shapes that accomplishes a desired behavior.

At the start of each step in a gait, all six motors are sent commands given by the PID controller.  The first time one tendon's length falls within the tolerance of the target length, that motor is stopped until the next step in the gait begins, which occurs once all six motors have reached their target (within tolerance) and have been stopped.

The gait used for most of the experiments is the robot's quasistatic (punctuated) rolling gait.  This gait consists of two steps: (1) a transition step which tips the robot onto the correct base triangle and (2) a rolling step where the robot deforms its body to shift its center of mass outside of the triangle of stability and roll onto the next face.  The transition step involves contracting two tendons to control the direction the robot tips, and the rolling step involves contracting the bottom two tendons to make the robot unstable and one other tendon to influence the direction of the roll.  The quasistatic rolling gait is defined as
$$[[1, 1, 0.1, 1, 1, 0.1], [0, 1, 1, 0, 1, 0.1]]$$
where each number is one actuator's target length, and a set of six is a target shape.  The target shape is specified in terms of the target lengths $[A, B, C, D, E, F]$ for each tendon as defined in Fig. S4 where the robot is in its rest state.  The specific tendons that are contracted change based on symmetry each time the robot rolls to a new face.  The full gait is given alongside the discussion of the robot's symmetry in SM section S6.  The robot's symmetry is illustrated in Fig.~\ref{fig:symmetry}.

The robot achieves its top locomotion speed of 18 body lengths per minute (BL/min) in its dynamic rolling given by $$[[0, 1, 1, 0, 1, 0.1]]$$ where it skips the transition step from the quasistatic gait and instead executes repeated rolling steps.  Skipping the transition step makes the rolling gait much faster (Fig.~\ref{fig:locomotion}K) at the expense of making it somewhat less reliable; on occasion after a step in the dynamic gait, the robot will fail to roll.  The gait parameters for the dynamic rolling experiments in Fig.~\ref{fig:locomotion}E are a minimum length of 100~mm, a range of 90~mm, a max speed of 99, a tolerance of 7\%, $K_P = 10$, $K_I = 0.01$, and $K_D = 0.5$.

While these two gaits were first discovered by hand, our simulation based on a differentiable physics engine~\cite{wang2022real2sim2real} generated a qualitatively similar rolling gait.  Moreover, the simulation generated turning gaits that were not discovered by hand.  The counterclockwise turning gait is $$[[1, 1, 1, 0, 1, 1], [1, 0, 1, 0, 1, 1], [0, 0, 0, 0, 0, 0], [1, 1, 1, 1, 1, 1]]$$ and the clockwise turning gait is $$[[0, 0, 0, 1, 0, 1], [0, 0, 0, 0, 0, 1], [0, 0, 0.8, 0, 1, 1], [1, 1, 1, 1, 1, 1]]$$ where the clockwise turning gait is extended via symmetry reduction as described in SM section S6.  The experiments for characterizing the turning gaits are described in SM section S3, and their results are shown in Fig.~\ref{fig:turning}.  These two turning gaits, along with the quasistatic rolling gait, are used in the trajectory following experiments in Fig.~\ref{fig:trajectory}, which are described in more detail below (section M4).

\subsubsection{Multi-terrain Experiments}

The multi-terrain experiments were conducted on two 24'' x 48'' patches of real grass (Fresh Patch), ice at the David S. Ingalls ice rink, polished pebbles (GASPRO), and play sand (Quikrete) as well as a flat floor covered by a white fabric background.  The experiments characterized the quasistatic rolling gait, and the gait parameters are given in Table S1. 

\begin{table}[htbp]
    \centering
    \begin{tabular}{|c|c|c|c|c|c|}
        \hline
        Terrain & Floor & Grass & Ice & Pebbles & Sand \\
        \hline
        Minimum Length (mm) & 100 & 100 & 100 & 100 & 100 \\
        \hline
        Range (mm) & 90 & 90 & 100 & 90 & 100 \\
        \hline
        Tolerance (\%) & 12 & 10 & 10 & 10 & 15 \\
        \hline
        Max Speed & 99 & 99 & 99 & 99 & 99 \\
        \hline
        $K_P$ & 8 & 6 & 6 & 6 & 6 \\
        \hline
        $K_I$ & 0.01 & 0.01 & 0.01 & 0.01 & 0.01 \\
        \hline
        $K_D$ & 0.5 & 0.5 & 0.5 & 0.5 & 0.5 \\
        \hline
    \end{tabular}
    \caption{Gait parameters for terrain experiments.}
    \label{tab:terrain_stats}
\end{table}

\subsubsection{Incline Experiments}

\begin{figure}
    \centering
    \makebox[\textwidth][c]{\includegraphics[width=7.24in]{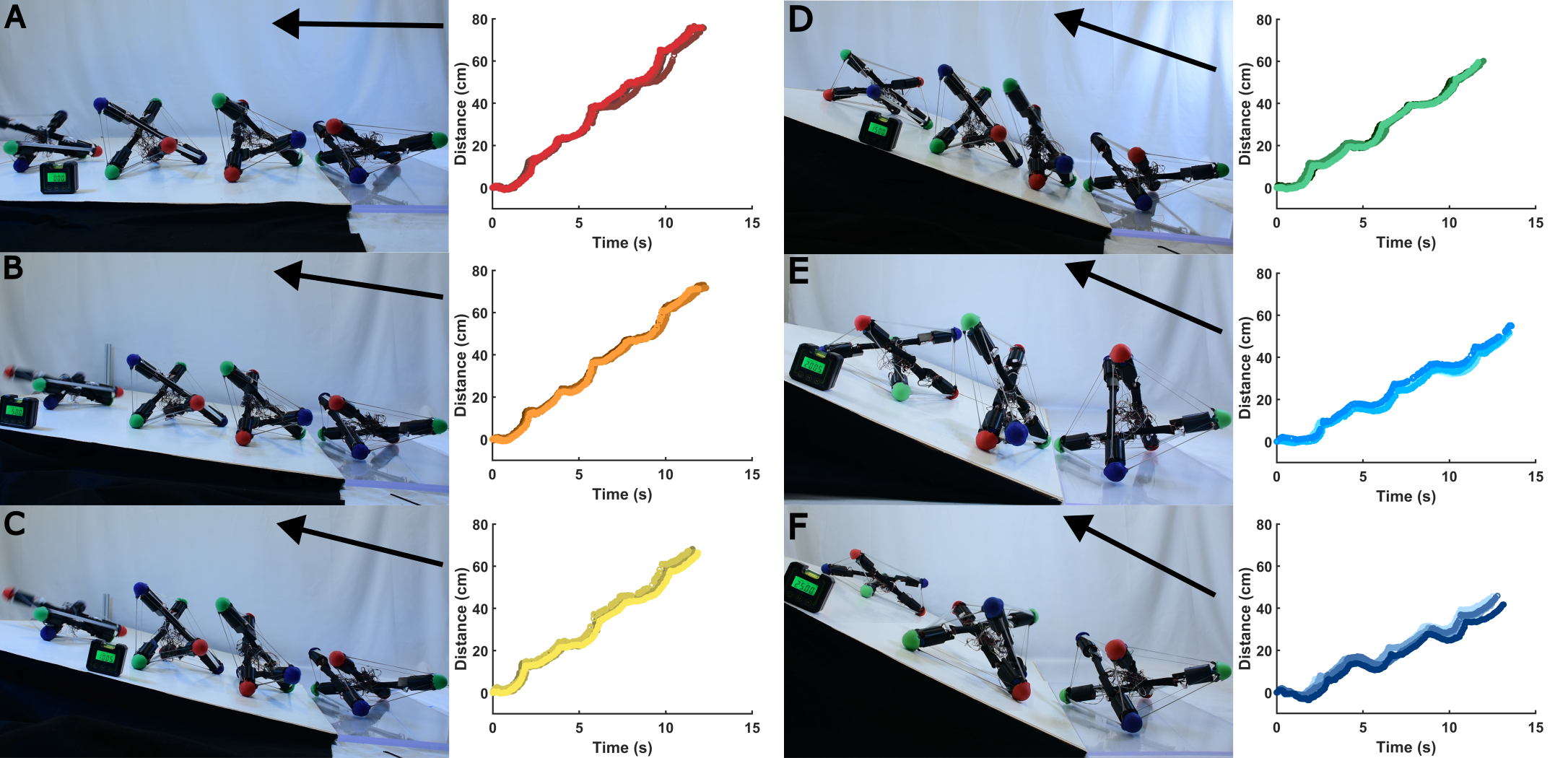}}
    \caption{\textbf{Incline experiments.} Photos and plots of incline experiments with angles of (A) 0\degree, (B) 5\degree, (C) 10\degree, (D) 15\degree, (E) 20\degree, and (F) 25\degree. Plots track the center of mass for three gait cycles over four trials. Gait parameters are given in Table~\ref{tab:incline_stats}.}
    \label{fig:inclines}
\end{figure}

The incline experiments were performed on two side-by-side 36'' x 24'' x 1/2'' sheets of medium density fiberboard (2726N42; McMaster-Carr) coated with an adhesive-backed silicone sheet and held at various incline angles.  The robot performed the quasistatic gait, and the gait parameters are given in Table~\ref{tab:incline_stats}.  To climb the steepest inclines, the gait's range needed to be high so that the robot could take bigger steps without slipping backward.  This change meant that the gait parameters had to be adjusted to avoid a problem with triangle inequality: at certain steps in the gait, two of the three tendons that make a triangle on one side of the robot are commanded to be short, close to their minimum length, while the third side is commanded to be long, forming a theoretically impossible triangle.  This problem is exacerbated by the increased shifting of the wires, which gradually changes the baseline capacitance of the sensor tendons, thereby adding noise, when the robot exhibits extreme deformations at this increased range.  Therefore, the tolerance parameter of the gait was split into {\em high tolerance}, which is the tolerance for target lengths closer to 1, and {\em low tolerance}, which is the tolerance for target lengths closer to 0.  The high tolerance was tuned to be higher than the low tolerance to accommodate the increased sensor noise and triangle inequality without compromising the critical mechanism of the gait: the small polygon of stability that allows the robot to roll over.  The experiments from inclines at various angles are shown in Fig.~\ref{fig:inclines} and movie S5.

\begin{table}[htbp]
    \centering
    \begin{tabular}{|c|c|c|c|c|c|c|c|}
        \hline
        Incline Angle & 0\degree & 5\degree & 10\degree & 15\degree & 20\degree & 25\degree & 28\degree \\
        \hline
        Minimum Length (mm) & 100 & 100 & 100 & 100 & 100 & 100 & 100 \\
        \hline
        Range (mm) & 140 & 140 & 140 & 140 & 140 & 140 & 180 \\
        \hline
        High Tolerance (\%) & 20 & 20 & 20 & 20 & 15 & 15 & 20 \\
        \hline
        Low Tolerance (\%) & 10 & 10 & 10 & 10 & 10 & 10 & 10 \\
        \hline
        Max Speed & 99 & 99 & 99 & 99 & 99 & 99 & 99 \\
        \hline
        $K_P$ & 10 & 10 & 10 & 10 & 10 & 10 & 10 \\
        \hline
        $K_I$ & 0.01 & 0.01 & 0.01 & 0.01 & 0.01 & 0.01 & 0.01 \\
        \hline
        $K_D$ & 0.5 & 0.5 & 0.5 & 0.5 & 0.5 & 0.5 & 0.5 \\
        \hline
    \end{tabular}
    \caption{Gait parameters for incline experiments.}
    \label{tab:incline_stats}
\end{table}

\subsection{Trajectory Following}
\label{sec:trajectory_following}
Shown in Fig. 4, the tensegrity robot can follow prescribed trajectories by selecting actions that minimize a cost function.  The actions include the quasistatic rolling gait with different values for the range (every 10~mm from 90~mm to 150~mm), the counterclockwise (ccw) turning gait with a range of 100~mm, the clockwise (cw) turning gait with a range of 100~mm, and modified versions of the quasistatic rolling gait where the range is different on either side of the robot.  When one side of the robot has a larger range than the other in the rolling gait, it takes a larger step on one side than the other, and the robot's behavior is forward motion with gradual turning (in contrast to the cw and ccw ``turn in place'' gaits) analogous to differential drive in wheeled robots when the wheels on one side spin faster than the other.  The following gait parameters are the same for all the actions: a minimum length of 100~mm, a max speed of 99, a tolerance of 20\%, $K_P = 6$, $K_I = 0.01$, and $K_D = 0.5$.

Each of these 51 actions was first modeled offline using our simulation based on a differentiable physics engine~\cite{wang2022real2sim2real}.  The robot's pose is the 2D position of the centroid of its six end caps and the 2D orientation of the robot given by the unit vector that points from the centroid of nodes 1, 3, and 5 to the centroid of nodes 0, 2, and 4 (see Fig.~\ref{fig:labels}), which we call the robot's principal axis.  Before and after executing each action in simulation, the pose (position and orientation) is extracted from the simulation.  The 2D rotation from the pose before to the pose after is calculated based on the angle by which the principal axis rotated about the z-axis.  The 2D translation from the pose before to the pose after is calculated relative to the robot's initial body frame via $t^0_1 = R^0_\mathcal{W} d^\mathcal{W}_1$ where
% $R_0^\mathcal{W} = \left[ \begin{matrix} x_0^\mathcal{W} & y_0^\mathcal{W} \end{matrix} \right]$
$R_0^\mathcal{W} = \left[ \begin{matrix} c_\theta & -s_\theta \\ s_\theta & c_\theta
\end{matrix}\right]$
is the rotation matrix that describes the robot's initial body frame with respect to the world frame given the anlge $\theta$ between the robot's principal axis and the world frame's x-axis.  The 2D rotation and translation relative to the robot's body frame before the action are stored in a data table, which is searched in real time during trajectory following trials to determine the best sequence of actions to minimize the cost function.

Each trajectory is represented as a series of 2D waypoints in a global frame.  An overhead RGB-D camera (Intel RealSense L515) was calibrated using April tags~\cite{richardson2013aprilcal} to measure its extrinsic matrix relative to the global frame.  Our previously developed tensegrity pose tracking algorithm~\cite{lu20226n} was modified to run in real time as a ROS service.  During each planning step, the node that controls the tensegrity robot requests the robot's current pose from the tracking service and plans a sequence of actions to follow the given trajectory.

The cost function is given by 
\begin{equation*}
    c = w_d \left|\left| \Vec{p}_i - \Vec{x} \right|\right| + w_a \left| \arccos \left( \Vec{h} \cdot \frac{\Vec{p}_{i+1} - \Vec{p}_i}{\left|\left| \Vec{p}_{i+1} - \Vec{p}_i \right|\right|} \right) \right| + w_p \left( 1 - \frac{i}{n} \right)
\end{equation*}
where $w_d$, $w_a$, and $w_p$ are the hand-tuned weights for distance, angle, and progression, respectively, $\Vec{p}_i$ is the 2D position vector of waypoint $i$ (the closest waypoint on the trajectory to the robot's current position), $\Vec{x}$ is the robot's 2D position given by the tracking service, $\Vec{h}$ is the robot's heading (the robot's principal axis rotated by 90\degree counterclockwise), and $n$ is the number of waypoints in the trajectory.  As in previous work~\cite{huang2022design,nagabandi2018learning}, the three terms in this cost function punish the robot for its 2D Euclidean distance from the trajectory, for the angle between the robot's heading and the trajectory's tangent at the robot's location, and for being closer to the start of the trajectory.  The weights used for the experiments are listed in Table~\ref{tab:trajectory_weights}.

The robot starts the trajectory following task by planning via an exhaustive tree search of all possible actions with a search depth of two.  For each action, the rotation and translation tabulated from the simulation are applied to the robot's current position and orientation to predict the next position and orientation after taking that action.
Then, for each prediction, we search the action space again, applying the transformation to each predicted position and orientation.  The resulting position $t^\mathcal{W}_2$ and orientation $R^\mathcal{W}_2$ with respect to the world frame after applying two actions are given by the equations below.
\begin{equation*}
    R^\mathcal{W}_2 = R^\mathcal{W}_0 R^0_1 R^2_1
\end{equation*}
\begin{equation*}
    t_2^\mathcal{W} = R_0^\mathcal{W} t_1^0 + R^\mathcal{W}_0 R_1^0 t_2^1
\end{equation*}
Having reached the prescribed search depth, the cost function is evaluated, and the sequence of two actions with the lowest cost is returned.  The robot executes the first action and then again plans a sequence of two actions, repeating this process until it reaches the end of the trajectory.

\begin{figure}
    \centering
    \makebox[\textwidth][c]{\includegraphics[width=7.24in]{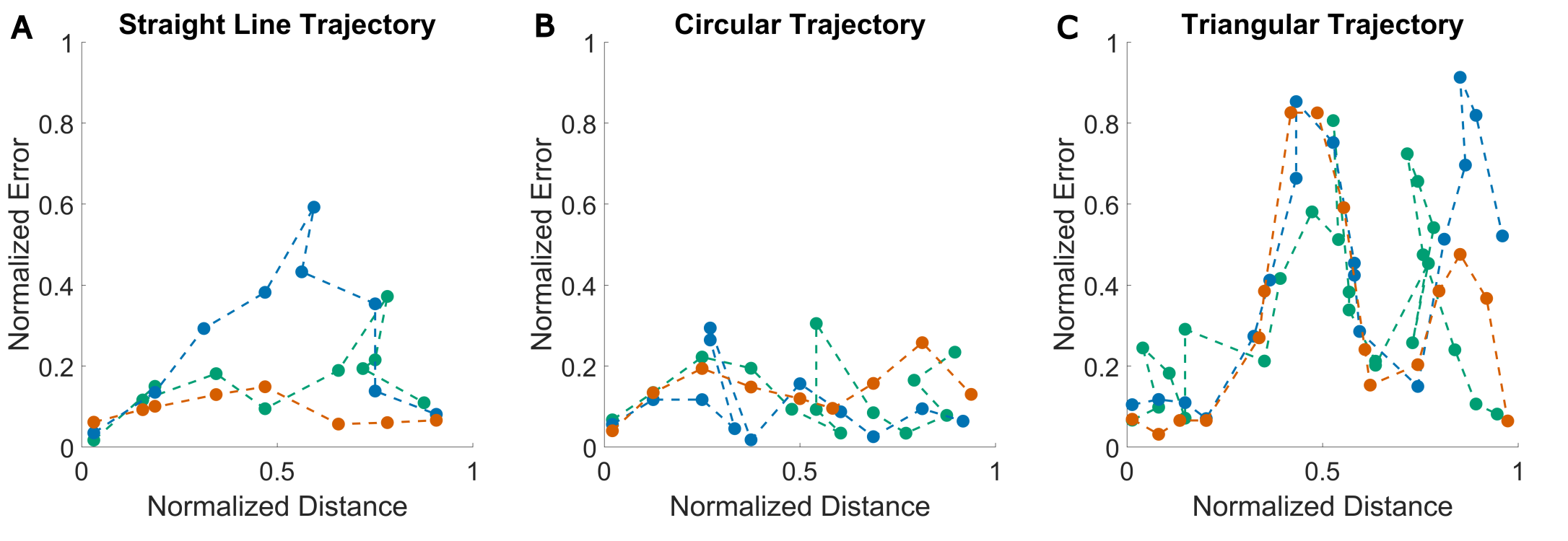}}
    \caption{\textbf{Trajectory following error.} Error as a function of distance along the trajectory for (A) straight line, (B) semicircle, and (C) right triangle trajectories.  The error is the shortest Euclidean distance from the robot's center of mass to a waypoint on the trajectory normalized by the robot's bar length.  The horizontal axis is the robot's distanced traveled along the trajectory normalized by the length of the trajectory.  The error is only evaluated during planning steps (markers) because those steps are when the cost function is evaluated.  The plots show three trials for each trajectory, and the colors correspond to Fig.~\ref{fig:trajectory}.}
    \label{fig:trajectory_error}
\end{figure}

Fig.~\ref{fig:trajectory} shows the robot's ability to follow a straight line, a circular arc, and a right triangle.  To follow the straight line, the controller usually commands rolling actions, course correcting for any deviations, and occasionally turn in place actions.  To follow the circle, gradual turning actions are selected.  The triangular trajectory's sharp corners pose further challenges.  To follow the triangle, the controller segments the triangle trajectory into three straight line trajectories.  When the robot reaches the end of a segment---defined as being within 27~cm of the last waypoint or if the last waypoint is the nearest point to the robot---the robot moves on to following the next segment.  At this juncture, to handle sharp corners, the controller decides if the robot should continue to roll in the same direction or instead, by virtue of its symmetry, roll backward by evaluating 
\begin{equation*}
    \Vec{a} \times \frac{\Vec{p}_{i+1} - \Vec{p}_i}{\left|\left| \Vec{p}_{i+1} - \Vec{p}_i \right|\right|} \geq 0
\end{equation*}
where $\Vec{a}$ is the robot's principal axis, defined above, and $\frac{\Vec{p}_{i+1} - \Vec{p}_i}{\left|\left| \Vec{p}_{i+1} - \Vec{p}_i \right|\right|}$ is the unit vector in the direction of the next trajectory segment.  If this inequality is true, the robot rolls in its typical forward direction.  Otherwise, if the cross product is negative, the robot rolls in the reverse direction, the gaits are modified via symmetry reduction, and the principal axis and heading are reversed in the robot body frame (SM section S6).  When the robot begins subsequent trajectory segments, it doesn't start with the perfect position and heading like it does for the first segment (and for the straight line and semicircle trajectories).  These errors compound, which is why the deviation from the trajectory is greater for the triangle than the other two.  However, the robot is still able to recover and reach the desired end point of the trajectory.  Plots of the error as a function of distance along the trajectory are shown in Fig.~\ref{fig:trajectory_error}.

\begin{table}[htbp]
    \centering
    \begin{tabular}{|c|c|c|c|}
        \hline
        Trajectory & $w_d$ & $w_a$ & $w_p$ \\
        \hline
        Straight Line & 500 & 200 & 300 \\
        \hline
        Circular Arc & 400 & 100 & 300 \\
        \hline
        Right Triangle & 600 & 100 & 300 \\
        \hline
    \end{tabular}
    \caption{Cost function weights for trajectory following experiments.}
    \label{tab:trajectory_weights}
\end{table}

\subsection{Limbo Demonstration}

To enable Tribar to play limbo autonomously, the pose tracking service was further augmented to perceive the height of the yellow limbo bar (SM section S4).  The robot first shrinks to the proper height, setting its range to the measured bar height minus 120~mm (100~mm accounting for the minimum length of the tendon and 20~mm for the diameter of the bar) with a maximum range of 140~mm and rounded down to the nearest 10~mm.  Then, the robot executes its quasistatic rolling gait at that range with a max speed of 99, a tolerance of 20\%, $K_P = 6$, $K_I = 0.01$, and $K_D = 0.5$.  When the robot has reached (within 27~cm) the endpoint of its 70-cm path under the limbo bar, as measured by the tracking service (SM section S4), the robot returns to its maximum height (at a range of 140~mm) and waits for the limbo bar to lower.  A stepper motor lowers the bar height by 40~mm; when the lowering of the bar is complete, the robot is manually commanded to resume the limbo demonstration with the newly lowered bar in the reverse direction, autonomously modifying its gait by symmetry reduction (SM section S6) and the gait's range based on the perceived bar height (SM section S4).

\subsection{Impact Resistance}

To survive the harsh impacts from rolling off the cliff (Fig.~\ref{fig:splash}) and the bridge (Fig.~\ref{fig:limbo_cliff}), the design of the end caps had to be modified.  As shown in Fig.~\ref{fig:endcaps}, the depth of the central hole through the end cap was increased so that the carbon fiber rod reinforced the stress concentration where the hemisphere meets the cylinder.  Also, the end caps were 3D printed from a stronger material (Tough 1500 Resin on a Form 3; Formlabs).
% The motors were stuck into the cylindrical mounts with a small amount of VHB (3M) to keep them from sliding out on impact.
The quasistatic rolling gait was modified to be $$[[1, 1, 0.1, 1, 1, 0.1], [0, 1, 1, 0, 1, 0.1], [1, 1, 1, 1, 1, 1]]$$ with the robot returning to its rest shape immediately after the rolling step because it is more capable of absorbing impact in that shape (i.e., more energy is absorbed through deformation before the rigid components contact each other).  For this gait, the minimum length is 100~mm, the range is 100~mm, the tolerance is 12\%, $K_P = 6$, $K_I = 0.01$, and $K_D = 0.5$.
% The results of the bridge dive experiment are shown in Figs. 1C and 5C as well as movie SX.

The cliff dive demonstration, shown in Fig.~\ref{fig:splash} and movie S1, is autonomous.  Sensor information is received over Bluetooth, and state estimation is calculated in real time.  The robot's initial heading is recorded based on the state estimation result at the beginning of the trial. 
 After each cycle of the modified quasistatic rolling gait, when the robot returns to its rest shape, the most recent state estimation result is analyzed to determine the robot's heading and which three nodes define the downward face.  The quasistatic gait is then modified with symmetry reduction (SM section S6), including reversing the gait if necessary, to ensure the gait is correct for the next cycle.

 The bridge dive demonstration, shown in Fig.~\ref{fig:limbo_cliff} and movie S8, uses the same gait, but the demonstration is not fully autonomous.  The state estimation algorithm relies on absolute orientations from the 9-axis IMUs, and the magnetic interference from the nearby buildings causes the magnetometer readings to drift.  For that reason, the robot is programmed to briefly pause after one gait cycle, and then the correct heading and downward face are manually entered depending on how the robot lands.  Afterward, the modified quasistatic rolling gait is executed continuously.

 In an additional field test, shown in movie S11, we demonstrate the robot rolling off a large cliff, about 12~m high, and recovering its shape at the bottom.  The height of the large cliff does not represent the greatest survivable drop height of the robot since it was not a shear drop: the robot was buffeted by the cliff side and vegetation on its way down.  This experiment was conducted for the lack of controls in an unpredictable, authentic field environment.  The drop was high enough that our Bluetooth communication was out of range, so the long delay in the video is due to the time it took to walk to the bottom of the cliff and reconnect to the robot.

\section*{Supplementary Text}
\renewcommand{\thesubsection}{S\arabic{subsection}}
\setcounter{subsection}{0}
\subsection{State of the Art of Tensegrity Robotics}

% talk about how tensegrity robots are progressing, but the grand challenge of autonomous conrol persists.  Here, we outline important metrics for our work (grand challenge #4), and we can only compare with what people report.  Point to reveiw papers for more comprehensive discussions.

Tensegrity robotics researchers have built many innovative robots and characterized their advantages of impact resistance, light weight, and versatility. As we work toward fully autonomous tensegrity robots that can navigate unstructured environments, it is important to measure the progress of our field with quantitative metrics~\cite{shah2021tensegrity}.  In this section, we report the critical performance metrics of Tribar that enable its key achievements of autonomy, versatility, and impact resistance and compare them to prominent examples of the state of the art.  This list is not meant to be exhaustive, in part because there are not universally agreed upon metrics for tensegrity robots that all studies report~\cite{shah2021tensegrity}.  A more comprehensive discussion of the state of the art can be found in recent review articles~\cite{shah2021tensegrity,liu2022review}.

% \begin{enumerate}
%     \item state estimation
%     \item multi-terrain
%     \item inclines
%     \item trajectory following
%     \item impact resistance
% \end{enumerate}

\subsubsection{State Estimation}

% elaborate on the main text discussion of prior art
State estimation from on-board sensors is a critical capability that enables Tribar's autonomous control in a field environment (Fig.~\ref{fig:splash}). Prior work on tensegrity state estimation has relied on off-board sensors~\cite{caluwaerts2016state} that will not realistically be available in the field environments that we envision future tensegrity robots will navigate.  Two previous studies present tensegrity state estimation with only on-board sensors.  Both use stretchable strain sensors: resistance-based conductive thread sensors~\cite{li2021shape} and capacitance-based silicone graphite sensors~\cite{booth2020surface}.  In both cases, the state estimation experiments are conducted in a stationary setting since the robots are limited to low strains.  In contrast, our highly stretchable sensor tendons accommodate the high strains necessary for Tribar's locomotion, so we conduct the state estimation experiments during locomotion.  Futhermore, the on-board IMUs enable Tribar to reconstruct not just its shape but also its orientation relative to Earth's gravitational and magnetic fields, a step beyond detecting only the bottom face~\cite{booth2020surface}.  The state of the art of tensegrity state estimation with on-board sensors is summarized in Table~\ref{tab:SOTAstateestimation}.  The root mean square error (RMSE) of state estimation is reported in absolute units and relative to the robots' respective bar lengths.

\begin{table}[h]
\centering
\begin{tabular}{|c|c|c|c|}
\hline
    Robot & Estimation Capabilities & Strain & RMSE \\
\hline
    Conductive Thread Tensegrity~\cite{li2021shape} & Shape & 63\% & 3.8~mm (1\%) \\
\hline
    Robotic Skins Tensegrity~\cite{booth2020surface} & Shape and downward face & 12\% & 45.8~mm (13.1\%) \\ % 
\hline 
    Tribar (ours) & Shape and global orientation & 180\% & 36~mm (10\%) \\
\hline
\end{tabular} \\
\caption{State estimation of tensegrity robots with on-board sensors.}
\label{tab:SOTAstateestimation}
\end{table} 

\subsubsection{Locomotion and Unstructured Terrain}

Tensegrity robots are motivated for their ability to adapt to unstructured terrain, yet most are only tested in controlled laboratory settings.  Some notable examples of tensegrity robots demonstrated on unstructured terrain include SUPERball demonstrated on NASA's roverscape~\cite{vespignani2018design}, a light-powered tensegrity with liquid crystal elastomer tendons demonstrated on sand and pebbles~\cite{wang2019light}, the TT-3 tensegrity demonstrated outdoors on dirt~\cite{chen2017soft}, and air-powered modular tensegrities that can crawl through a dirt tunnel~\cite{kobayashi2022soft}.  Aside from terrain, another challenge with tensegrity locomotion is that it is typically slow for mobile robots.  Table~\ref{tab:SOTA_speed_terrain} lists the characteristic velocity in bar lengths per minute (BL/min) for some of the fastest tensegrity robots in the literature as well as whether they have been demonstrated on unstructured terrain.  The curved bar tensegrity is capable of pure rolling, not just the punctuated rolling that is typical for tensegrity robots. 
 The pure rolling and vibrating robots exhibit faster speeds than Tribar's punctuated rolling; however, Tribar is not limited to locomotion on flat planes in a laboratory setting.

% write a little bit about the top speed robots (point to table) and other slow examples of unstructured terrain (ryota, LCE, etc.)

\begin{table}[h]
\centering
\begin{tabular}{|c|c|c|c|}
\hline
    Robot & Characteristic Velocity (BL/min) & Unstructured Terrain \\
\hline 
    TT-4\textsubscript{mini}~\cite{chen2017inclined} & 11 & - \\
\hline
    Curved Bar Tensegrity~\cite{kaufhold2017indoor} & 24 & - \\
\hline
    Vibrating Tensegrity~\cite{rieffel2018adaptive} & 69 & - \\
\hline                                                            
    SUPERball~\cite{vespignani2018design} & 3.5 & NASA Roverscape \\
\hline 
    Tribar (ours) & 18 & grass, ice, pebbles, sand, cliffs \\
\hline
\end{tabular} \\
\caption{Locomotion of tensegrity robots.}
\label{tab:SOTA_speed_terrain}
\end{table} 

\subsubsection{Maximum Climbable Incline}

Table~\ref{tab:SOTAincline} summarizes the inclined locomotion capabilities of tensegrity robots with gravity-based rolling gaits.  Some studies also demonstrate stable tensegrity locomotion down declines.  We exclude crawling-based gaits---like those of pipe-climbing tensegrity robots that are capable of vertical locomotion~\cite{friesen2016second,kobayashi2022soft}---as well as wheeled tensegrity robots~\cite{spiegel2023shape} from this comparison.  Tribar's locomotion up a 28\degree incline is the steepest incline climbed by a tensegrity robot in the reported literature.

\begin{table}[h]
\centering
\begin{tabular}{|c|c|c|}
\hline
    Robot & Maximum Climbable Incline & Maximum Stable Decline \\
\hline
    Robotic Skins Tensegrity~\cite{booth2020surface} & - & 8.7\degree \\
\hline
    Curved Bar Tensegrity~\cite{kaufhold2017indoor} & 7\degree & 7\degree \\
\hline                                                            
    SUPERball~\cite{surovik2021adaptive} & 15\degree & - \\
\hline 
    TT-4\textsubscript{mini}~\cite{chen2017inclined} & 24\degree & - \\
\hline 
    Tribar (ours) & 28\degree & 20\degree \\
\hline
\end{tabular} \\
\caption{Inclined locomotion of tensegrity robots.}
\label{tab:SOTAincline}
\end{table} 

\subsubsection{Drop Height}

Only a few studies on tensegrity robots experimentally demonstrate impact resistance by showing the robots surviving a drop, and these demonstrations are summarized in Table~\ref{tab:SOTAimpact}.  NASA's SUPERBall, a 6-bar tensegrity robot with a diameter of 2~m, was shown falling 3.4~m from the top of a roof~\cite{vespignani2018design}.  Rieffel dropped his vibrating tensegrity robots from his eye level~\cite{rieffel2018adaptive}.  Another team of researchers built a drone with tensegrity landing gear and a rover with a tensegrity axle, which survived drops of 2~m and 5~m, respectively~\cite{zappetti2022dual}, and yet another group built a robot with tensegrity wheels~\cite{spiegel2023shape}. Even though these robots are not pure tensegrity structures like SUPERball, Tribar, and the vibrating tensegrities, we include them as examples in Table~\ref{tab:SOTAimpact} because they still demonstrate how incorporating tensegrities into robots enables impact resistance.  We have not included studies of the impact resistance of tensegrity structures that are not robots.  Tribar's 5.7-m fall from the bridge, described above, is the highest absolute drop survived by a tensgrity robot in the reported literature.  Normalizing by bar length yields higher normalized drop heights for smaller, lighter robots.

\begin{table}[h]
\centering
\begin{tabular}{|c|c|c|c|}
\hline
    Robot & Bar Length (m) & Drop Height (m) & Drop Height (BL) \\
\hline
    Tensegrity Wheels~\cite{spiegel2023shape} & 0.13 & 1.5 & 11.5 \\
\hline
    Vibrating Tensegrity~\cite{rieffel2018adaptive} & 0.094 & \~1.8 & \~19 \\
\hline
    Tensegrity Drone~\cite{zappetti2022dual} & 0.094 & 2 & 21 \\
\hline                                                            
    SUPERball~\cite{vespignani2018design} & 2 & 3.4 & 1.7 \\
\hline 
    Tensegrity Rover~\cite{zappetti2022dual} & 0.094 & 5 & 53 \\
\hline 
    Tribar (ours) & 0.36 & 5.7 & 15.8 \\
\hline
\end{tabular} \\
\caption{Drop heights survived by tensegrity robots.}
\label{tab:SOTAimpact}
\end{table} 

\subsubsection{Standard Metrics}

In a recent review paper~\cite{shah2021tensegrity}, we suggested uniform metrics that all tensegrity robotics papers should report to track the progress of the field.  Those metrics for our 3-bar tensegrity robot are listed in Table~\ref{tab:metrics}.  The robot density is defined as the robot's mass of 0.4~kg divided by its maximum volume in its most expanded state.  The characteristic rod length is 360~mm for all of the experiments except the bridge and cliff dive experiments, as described above.  The maximum bending, buckling, and crushing loads of the rods were measured in an Instron 3345 materials testing system using the nominal 360~mm rods (not the slightly shorter rods with the Tough 1500 end caps that were used in the impact tests).  The maximum buckling and crushing loads were measured in an axial compression experiment while the maximum bending load was measured in a 3-point bending experiment. The rate for these experiments was 1~mm/s.  The characteristic velocity is the robot's top speed during its dynamic rolling gait, described above, divided by its characteristic rod length.  The only standard metric we were unable to measure is the cost of transport because the robot does not have an on-board power sensor.  We plan to measure the cost of transport of Tribar in future work.

\begin{table}[h]
\centering
\begin{tabular}{|c|c|}
\hline
    Metric & Value \\
\hline
    Robot density ($kg/m^3$) &  56 \\
\hline 
    Characteristic Rod Length ($m$) & 0.36 \\
\hline                                                            
    Maximum Bending Load ($N$) & 1.3 \\
\hline 
    Maximum Buckling Load ($N$) & 1.2 \\
\hline                                                              
    Maximum Crushing Load ($N$) & 1.4 \\
\hline
    Characteristic velocity (BL/min) & 18 \\
\hline
    Maximum climbable incline & 28\degree \\
\hline
\end{tabular} \\
\caption{Standard metrics for our 3-bar tensegrity.}
\label{tab:metrics}
\end{table} 

\subsection{Shape Morphing}

\begin{figure}
    \centering
    \makebox[\textwidth][c]{\includegraphics[width=7.24in]{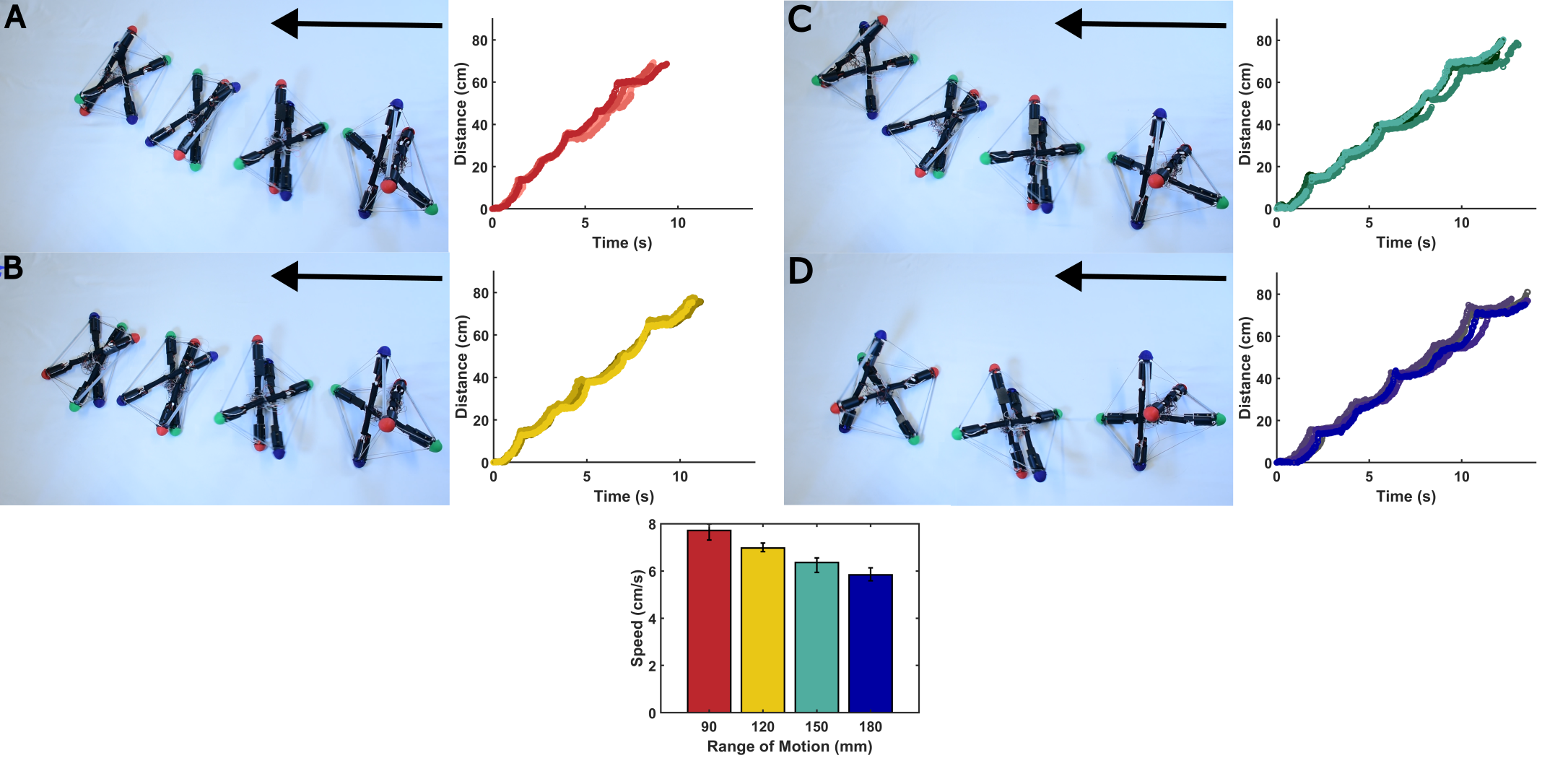}}
    \caption{\textbf{Shape morphing experiments.} Locomotion photos and plots for gaits with a varying range of (A) 90~mm, (B) 120~mm, (C) 150~mm, and (D) 180~mm. Plots track the center of mass for three gait cycles over four trials. Gait parameters are given in Table~\ref{tab:shape_morphing_stats}. The bar plot shows the average speed over four trials while error bars represent the maximum and minimum speeds.}
    \label{fig:shape_morphing}
\end{figure}

To understand the effect of the robot's shape morphing on its locomotion, we characterized the quasistatic rolling gait with a varying range (90~mm to 180~mm).  The experiments are shown in Fig.~\ref{fig:shape_morphing}, and the gait parameters are given in Table~\ref{tab:shape_morphing_stats}.  The robot's velocity decreases as the range increases.  Although a larger range results in bigger steps each time the robot rolls, contracting a longer cable takes more time for the motors, resulting in a lower frequency of rolls and a slower overall speed.

\begin{table}[htbp]
    \centering
    \begin{tabular}{|c|c|c|c|c|}
        \hline
        \textbf{Range (mm)} & \textbf{90} & \textbf{120} & \textbf{150} & \textbf{180} \\
        \hline
        Minimum Length (mm) & 100 & 100 & 100 & 100 \\
        \hline
        Tolerance (\%) & 12 & 12 & 15 & 15 \\
        \hline
        Max Speed & 99 & 99 & 99 & 99 \\
        \hline
        $K_P$ & 8 & 8 & 8 & 8 \\
        \hline
        $K_I$ & 0.01 & 0.01 & 0.01 & 0.01 \\
        \hline
        $K_D$ & 0.5 & 0.5 & 0.5 & 0.5 \\
        \hline
    \end{tabular}
    \caption{Gait parameters for shape morphing experiments.}
    \label{tab:shape_morphing_stats}
\end{table}

\subsection{Turning Experiments}

In addition to rolling forward, we demonstrate different turning gaits for the tensegrity robot.  The two most reliable gaits for turning in place were discovered by our simulator (SM section S5).  The counterclockwise turning gait is $[[1, 1, 1, 0, 1, 1], [1, 0, 1, 0, 1, 1], [0, 0, 0, 0, 0, 0], [1, 1, 1, 1, 1, 1]]$, and the clockwise turning gait is $[[0, 0, 0, 1, 0, 1], [0, 0, 0, 0, 0, 1], [0, 0, 0.8, 0, 1, 1], [1, 1, 1, 1, 1, 1]]$.  In the counterclockwise gait, the robot shifts its weight during the cycle to rotate its body, maintaining the same nodes on the bottom for the duration of the gait.  In the clockwise gait, like the rolling gaits, the robot flips over onto a new face.  Therefore, for the clockwise gait to be repeatedly executed, it is extended by symmetry reduction as described below.  The results of the experiments with these ``turn in place'' gaits are shown in Fig. 3, and the gait parameters are listed in Table~\ref{tab:turn_stats}.

\begin{figure}
    \centering
    \makebox[\textwidth][c]{\includegraphics[width=4.24in]{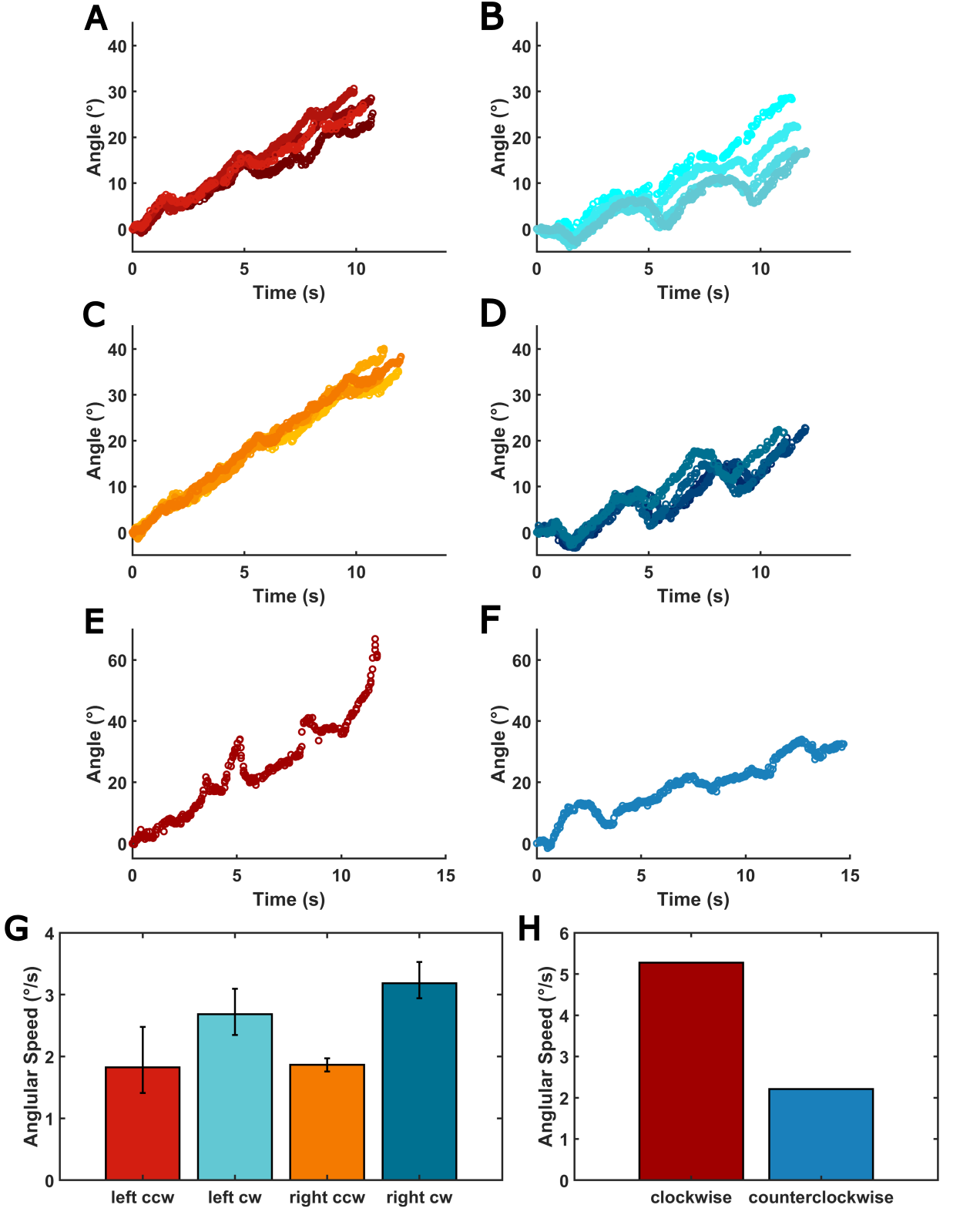}}
    \caption{\textbf{Turning experiments.} Locomotion plots for turning (A, C, E) ccw and (B, D, F) cw.  (A-D) Crawling-based turning gaits using (A-B) the left side or (C-D) the right side of the robot as a pivot. Plots track the center of mass for three gait cycles over four trials. (E-F) One trial each for gradual turning gaits. (G) A comparison of crawling-based turning speeds. (H) A speed comparison for gradual turning gaits.  The bar plots show the average speed while error bars represent the maximum and minimum speeds over the trials. Gait parameters are given in Table~\ref{tab:turn_stats}, and these gaits are shown in Movie S10.}
    \label{fig:turning}
\end{figure}

The robot is also capable of a crawling-based turning gait where it uses one side of its body as a pivot while the other side crawls along using friction.  Experiments, shown in Fig.~\ref{fig:turning}, characterize the robot turning clockwise and counterclockwise while using each side as the pivot.  Here, ``left'' refers to using the left side of the robot (with nodes 1, 3, and 5) as the pivot while ``right'' refers to using the right side (with nodes 0, 2, and 4) as the pivot.  ``Left'' and ``right'' are defined with respect to the robot's locomotion direction during the quasistatic rolling gait from its rest state (Fig.~\ref{fig:labels}).  The gait for crawling-based turning with the left pivot is given by $$[[0, 0, 0, 0.1, 0.1, 0.1], [0, 0, 0, 1, 0.1, 1], [0, 0, 0, 1, 1, 0.1]]$$ for turning clockwise and $$[[0, 0, 0, 0.1, 0.1, 0.1], [0, 0, 0, 1, 1, 0.1], [0, 0, 0, 1, 0.1, 1]]$$ for turning counterclockwise.  The gait can be adapted to use the right pivot by symmetry reduction as described below.  The gait parameters for these experiments are listed in Table~\ref{tab:turn_stats}.

Finally, the robot can also turn using the quasistatic rolling gait by using a different value for the range on each side of its body, analogous to differential drive on wheeled vehicles.  Just as spinning the left and right wheels at different speeds leads to a gradual turning motion on a wheeled robot, our robot can turn while rolling forward.  Experiments (Fig.~\ref{fig:turning}) show the robot gradually turning counterclockwise by employing the quasistaic rolling gait with a range of 80~mm on the left side and 160~mm on the right side.  Then, it can turn clockwise using the same gait by switching the range values.  Gait parameters for these gradual turning experiments are listed in Table~\ref{tab:turn_stats}.  This gradual turning gait and its variations are used in the control strategy for the trajectory following task shown in Fig.~\ref{fig:trajectory}.

\begin{table}[htbp]
    \centering
    \begin{tabular}{|c|c|c|c|}
        \hline
        Direction & Turn in place & Crawling turn & Gradual turning \\
        \hline
        Minimum Length (mm) & 100 & 100 & 100 \\
        \hline
        Range (mm) & 100 & 90 & 80/160 \\
        \hline
        Tolerance (\%) & 15 & 17 & 10 \\
        \hline
        Max Speed & 80 & 99 & 99 \\
        \hline
        $K_P$ & 6 & 6 & 6 \\
        \hline
        $K_I$ & 0.01 & 0.01 & 0.01 \\
        \hline
        $K_D$ & 0.5 & 0.5 & 0.5 \\
        \hline
    \end{tabular}
    \caption{Gait parameters for turning experiments.}
    \label{tab:turn_stats}
\end{table}

\subsection{Pose Tracking}

All the plots in this paper that track the robot's CoM are produced using a tensegrity pose tracking algorithm we previously published~\cite{lu20226n}.  This pose tracking algorithm also provides the ground truth for evaluating our state estimation experiments (Fig.~\ref{fig:state_recon} and Fig.~\ref{fig:state_reconstruction2}).  For the trajectory following experiments and the limbo demonstration, the algorithm was modified to run as a ROS service so it could be used online.

The pose tracking algorithm fuses sensor inputs from an RGB-D camera (Intel RealSense L515) and the on-board capacitive strain sensors described in section M1.3.  The algorithm tracks the pose of the three tensegrity rods by observing the red, green, and blue colors of the end caps in the camera data.  Information from the on-board sensors is weighted more by the algorithm when the end caps are heavily occluded, usually due to self-occlusions with the other rods.  We manually give the algorithm the initial positions of the end caps by clicking on them in numerical order (0-5 in Fig.~\ref{fig:labels}) on an image of the first frame of the RGB-D video.  The pose tracking for all the locomotion trials was done offline so we could plot the CoM and calculate the speed.

For the trajectory following experiments and the limbo demonstration, the pose tracking service was run online in real time.  The pose estimate was updated whenever the service received synchronized data from the ROS nodes publishing the RGB-D data and the robot's on-board sensor data.  After each gait cycle, the node controlling the robot requested the robot's current pose from the tracking service.  The robot's pose was used to modify the next cycle of the gait appropriately based on the bottom face of the robot (SM section S6).  The pose was also used for motion planning in the trajectory following experiments and to determine if the robot had gone past its endpoint on the other side of the bar in the limbo demonstration.

For the limbo demonstration, the pose tracking service is further augmented to perceive the height of the yellow bar.  Each time the robot begins a segment of limbo, the robot control node sends a request to the tracking service for the bar height.  The tracking service filters the latest RGB-D data for the yellow color of the bar and estimates the height with RANSAC.
% ~\cite{fischler1981random}.  
After the bar height is received from the tracking service, the robot shrinks to the appropriate size and adjusts the range of its gait (section M5) before beginning the limbo segment.  All the pose tracking code can be found in the GitHub repository linked in our previous work~\cite{lu20226n}.

\subsection{Simulation}

The simulation used in this work is a differentiable physics engine that we previously published~\cite{wang2022real2sim2real}.  The simulation was used to generate the clockwise and counterclockwise turning gaits discussed in section M3.2 and characterized in Fig.~\ref{fig:locomotion}.  It was also used to model all of the possible actions for the trajectory following experiments discussed in section M4 and shown in Fig.~\ref{fig:trajectory}.

The clockwise and counterclockwise gaits were generated using a graph search with the principal axis rotation as the reward function~\cite{wang2022real2sim2real}.  To simplify the search, the targets lengths were constrained to either fully extended or fully contracted (i.e., $1$ or $0$).  The gaits generated in simulation transferred to the real robot with only one modification to the third step in the clockwise turning gait.  The simulation output was $[0,0,1,0,1,1]$, but for the gait to execute more reliably, it was changed to $[0,0,0.8,0,1,1]$.  To achieve $[0,0,1,0,1,1]$, tendons A and B (see Fig.~\ref{fig:labels}) would go to the minimum length (e.g., 100~mm) and tendon C would have to reach full extension (e.g., 200~mm).  Depending on the range and tolerance used, these commands could produce a triangle inequality, which the robot could not possibly achieve.  The modification, changing C's target from $1$ to $0.8$, is designed to mitigate this issue.

The Real2Sim2Real pipeline we use with our differentiable simulator involves recording a trajectory on the real robot in order to perform system identification (sysID).  Therefore, prior to the trajectory following experiments, we recorded a simple, open-loop locomotion trajectory on the substrate where the experiments would be conducted (vinyl dance floor; VEVOR) to identify the friction parameters for the robot on this substrate.  More details about the sysID process can be found in our previous work~\cite{wang2022real2sim2real}.  After sysID, we modeled the robot's full action space in the simulator and recorded the robot's 2D transformations in a data table as described in section M4.  The data table used for the trajectory following task is available in the supplementary data.

\subsection{Symmetry Reduction}

\begin{figure}
    \centering
    \makebox[\textwidth][c]{\includegraphics[width=7.24in]{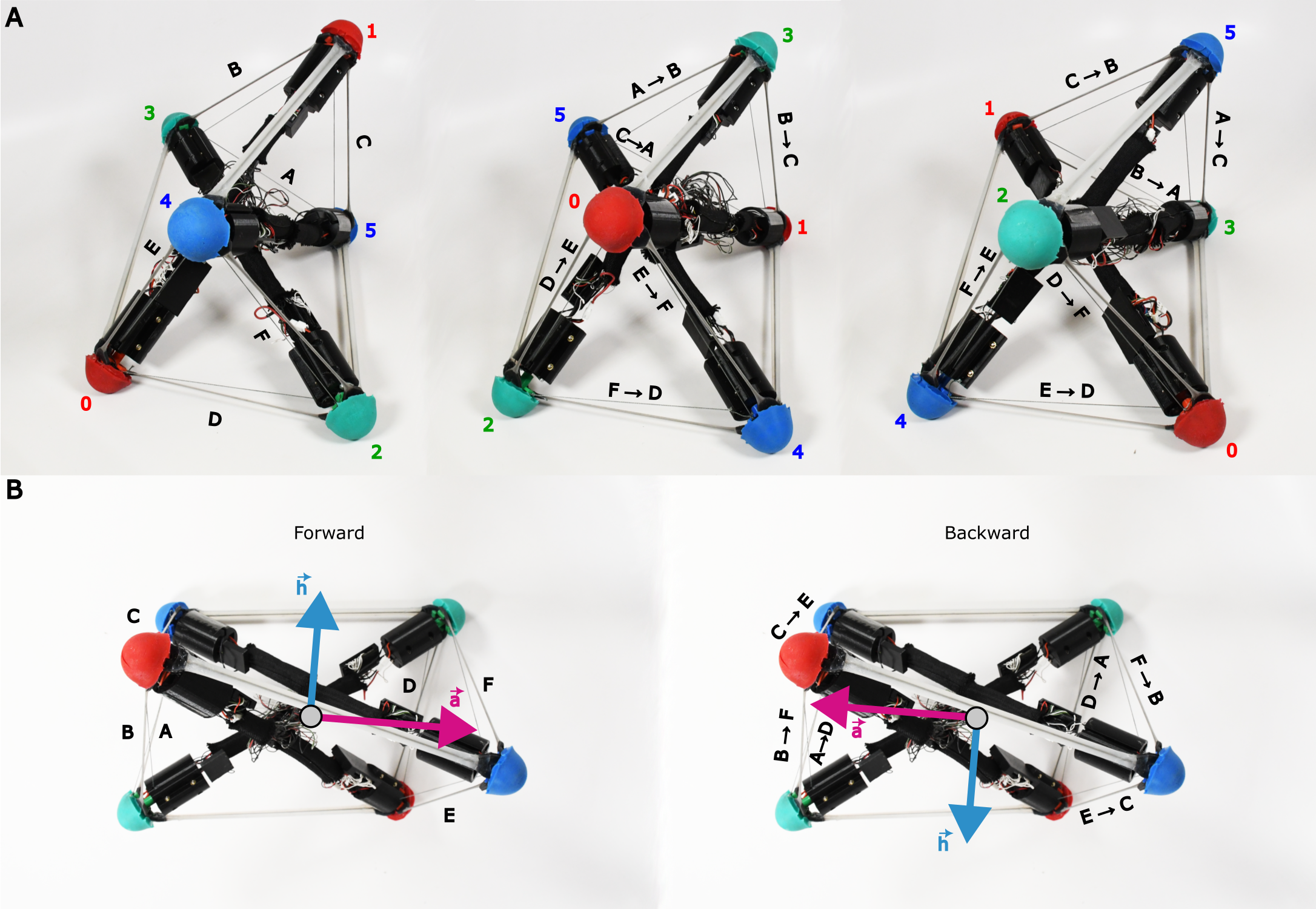}}
    \caption{\textbf{Robot Symmetry.} For each bottom face, there is a mapping of the actuator commands based on the robot's symmetry so that the robot can execute the same gait as it would from the rest state (when the bottom four nodes are 0, 2, 3, and 5).  There is also a mapping for reversing a gait, for example, to roll backward.  (A) The robot's gaits can be translated for each bottom face.  The robot is pictured on its three bottom faces, starting with the rest state, in the order it would land on them while executing the quasistatic rolling gait. For each bottom face, the gait mapping is labeled.  For example, in the middle image when nodes 1, 2, 4, and 5 are the bottom four, ``A $\rightarrow$ B'' means that actuator A (between nodes 3 and 5) should be given the target lengths that would be given to actuator B if the robot were in the rest state. (B) The mapping to reverse a gait can be visualized from the topdown view by rotating the robot 180\degree in the ground plane.  The bottom two tendons swap target lengths in the mapping, and the other four swap across-diagonally.  For this reason, the reverse mapping is a function of the bottom face.  When the quasistatic rolling gait is reversed, the robot rolls in the opposite direction, so for the purposes of the trajectory following experiments, the heading and principal axis are also reversed.  When the clockwise and counterclockwise turning gaits are reversed, they still yield clockwise and counterclockwise turning, respectively.}
    \label{fig:symmetry}
\end{figure}

Throughout this manuscript, the robot's locomotion gaits have been given based on its initial position with nodes 0, 2, and 5 touching the ground.  However, when another face is downward, the gaits can be adapted using the robot's symmetry.  For example, the quasistatic rolling gait is given as $[[1, 1, 0.1, 1, 1, 0.1], [0, 1, 1, 0, 1, 0.1]]$ for when nodes 0, 2, and 5 are on the ground.  In the second step---the rolling step---of the gait, the bottom two tendons (A and D) are fully contracted at 0, shrinking the polygon of stability to allow a roll.  After one gait cycle where the robot rolls over, nodes 1, 2, and 5 will be touching the ground, and tendons C and F will be the bottom two.  Therefore, the next two steps in the gait should be $[[1, 0.1, 1, 1, 0.1, 1], [1, 1, 0, 1, 0.1, 0]]$.  The full quasistatic rolling gait is therefore $$[[1, 1, 0.1, 1, 1, 0.1], [0, 1, 1, 0, 1, 0.1],$$ $$[1, 0.1, 1, 1, 0.1, 1], [1, 1, 0, 1, 0.1, 0],$$ $$[0.1, 1, 1, 0.1, 1, 1], [1, 0, 1, 0.1, 0, 1]]$$ after which the robot has the same orientation since three rolls have been completed, and therefore this six-step gait can be repeated from the beginning.

In this way, any gait can be adapted to a different bottom face.  This mapping is shown in Fig.~\ref{fig:symmetry}A.  As the robot rolls to the next face, the target lengths on each side (ABC vs. DEF) should be left-shifted by one.  For example, if the gait is $[0,1,2,3,4,5]$ when the bottom nodes are 0, 2, 3, and 5, then that gait should be translated to $[1,2,0,4,5,3]$ when the robot is on the next face with nodes 1, 2, 4 and 5 closest to the ground.  For the last face (nodes 0, 1, 3 and 4), the same gait would be $[2,0,1,5,3,4]$.

Using the same strategy, we can extend the clockwise turning gait from above, adapting it to the new faces in sequence after the robot rolls over.  Therefore, the full clockwise turn in place gait (starting with nodes 0, 2, and 5 on the ground) is $$[[0, 0, 0, 1, 0, 1],[0, 0, 0, 0, 0, 1],[0, 0, 0.8, 0, 1, 1],[1, 1, 1, 1, 1, 1],$$ $$[0, 0, 0, 1, 1, 0],[0, 0, 0, 1, 0, 0],[0.8,0,0,1,0,1],[1,1,1,1,1,1],$$ $$[0,0,0,0,1,1],[0,0,0,0,1,0],[0,0.8,0,1,1,0],[1,1,1,1,1,1]]$$ after which the robot has rolled back onto its original face.  Note that the clockwise gait rolls onto new faces in the opposite sequence compared to the quasistatic rolling gait, which is why the target lengths are right-shifted instead of left-shifted.  The counterclockwise turning gait does not cause the robot to roll onto a new face, but it can still be adapted to different bottom faces via the same strategy.

The quasistatic rolling gait can also be modified so that the robot rolls in the opposite direction (Fig.~\ref{fig:symmetry}B).  The full gait to roll backward is $$[[1,0.1,1,1,0.1,1],[0,0.1,1,0,1,1],$$ $$[1,1,0.1,1,1,0.1],[1,0,0.1,1,0,1],$$ $$[0.1,1,1,0.1,1,1],[0.1,1,0,1,1,0]]$$ when starting with nodes 0, 2, 3, and 5 closest to the ground.  The general mapping to reverse the quasistatic rolling gait is a function of the bottom nodes; the mapping can be found in the supplementary code.

The crawling-based turning gaits can be adapted to use the other side of the robot as the pivot foot.  Using the right pivot, the crawling-based turning gaits are $$[[0.1,0.1,0.1,0,0,0],[1,1,0.1,0,0,0],[0.1,1,1,0,0,0]]$$ for turning clockwise and $$[[0.1,0.1,0.1,0,0,0],[0.1,1,1,0,0,0],[1,1,0.1,0,0,0]]$$ for turning counterclockwise.  This reverse mapping for the crawling-based turning gaits is the same mapping for reversing the quasistatic rolling gait.  Notably, the clockwise and counterclockwise turn in place gaits can also be reversed, and they still yield clockwise and counterclockwise turning, respectively.

% \noindent {\bf Fig. 1.} Please do not use figure environments to set
% up your figures in the final (post-peer-review) draft, do not include graphics in your
% source code, and do not cite figures in the text using \LaTeX\
% \verb+\ref+ commands.  Instead, simply refer to the figure numbers in
% the text per {\it Science\/} style, and include the list of captions at
% the end of the document, coded as ordinary paragraphs as shown in the
% \texttt{scifile.tex} template file.  Your actual figure files should
% be submitted separately.

\end{document}